%% file: neurips_2026.tex
\documentclass{article}

\PassOptionsToPackage{numbers, compress}{natbib}

\usepackage[preprint]{neurips_2026}
\usepackage[utf8]{inputenc} 
\usepackage[T1]{fontenc}    
\usepackage{hyperref}       
\usepackage{url}            
\usepackage{booktabs}       
\usepackage{amsmath}        
\usepackage{amsfonts}       
\usepackage{graphicx}       
\usepackage{placeins}       
\usepackage{nicefrac}       
\usepackage{microtype}      
\usepackage[table]{xcolor}  
\usepackage{pifont}         
\usepackage{algorithm}      
\usepackage{algpseudocode}  

\definecolor{civablue}{RGB}{31,78,120}      
\definecolor{civarow}{RGB}{255,236,179}     
\definecolor{civastripe}{RGB}{240,244,248}  
\definecolor{civarule}{RGB}{31,78,120}      

\title{CIVA: Critic-Induced Value-Subspace Attacks on Visual World-Model Agents}

\author{
  Jiancheng Wang \quad
  Mingli Zhu \quad
  Tong Zhang \quad
  Jiaqi Ruan
  \\
  Wei Wang \quad
  Siyuan Liang$^{*}$ \quad
  Dacheng Tao
  \\[4pt]
  Corresponding author:\texttt{pandaliang521@gmail.com}
}

\begin{document}

\maketitle

\begin{abstract}
  Visual world-model agents such as DreamerV3 act through a recurrent latent state rather than a single observation, which weakens frame-wise observation attacks and makes their perturbations vary sharply over time under a strict per-frame perturbation constraint. We study white-box, causal, online attacks on such agents and propose Critic-Induced Value-Subspace Attacks (\textbf{CIVA}). Our key observation is that, along a rollout, critic-guided perturbations concentrate in a low-dimensional subspace induced by the victim's own critic. Based on this observation, CIVA first probes the frozen victim offline with critic-guided PGD and extracts a low-rank value-subspace by SVD. At test time, it optimizes only the subspace coefficients, smooths them with an exponential moving average (EMA), and maps them back to pixels. This design attacks value-sensitive recurrent dynamics while keeping the online optimization cheap and temporally coherent. Extensive experiments on DMC walker walk, Atari Pong, and Crafter show that CIVA consistently outperforms five recent methods; on DMC walker walk, it achieves the largest reward drop of 26.07\% while keeping temporal variation low, with TempAbs of 0.646.
\end{abstract}

\section{Introduction}
\label{sec:intro}
\input{Sec/1-intro.tex}

\section{Related Work}
\input{Sec/2-related-work.tex}

\section{Threat Model}
\input{Sec/3-threat-model.tex}

\section{Method}
\label{sec:method}
\input{Sec/4-method.tex}

\section{Experiment}
\input{Sec/5-experiment.tex}

\section{Conclusion}
\label{sec:conclusion}
\input{Sec/6-conclusion.tex}

\begin{ack}
\end{ack}

\medskip

{
\small
\bibliographystyle{plainnat}
\bibliography{main}
}


\appendix

\section{Supplementary Material}
\input{Sec/7-supplementary-material.tex}


\newpage

\end{document}

%% file: Sec/1-intro.tex
Visual world models are becoming a mainstream paradigm for visual decision-making. Rather than reacting to the current frame alone, Dreamer-style agents write each observation into a recurrent latent state and act through learned dynamics in that internal space~\citep{ha2018worldmodels,hafner2019dreamerv1,hafner2020dreamerv2,hafner2023mastering}. Related lines on planning with learned models~\citep{schrittwieser2020muzero}, transformer-~\citep{micheli2023iris,robine2023twm,zhang2023storm} and diffusion-based world models~\citep{alonso2024diamond}, large-scale generative environments~\citep{bruce2024genie}, scalable model-based control~\citep{hansen2024tdmpc2}, and robot deployment~\citep{wu2023daydreamer} further reinforce recurrent latent dynamics as a core primitive in visual RL~\citep{ding2024worldmodelsurvey,guo2026world,xu2026ctrlattack}. As these agents move closer to safety-critical use, understanding how to attack them becomes increasingly important.

Existing adversarial attacks on RL already show that deep policies are highly sensitive to small perturbations in visual or state observations~\citep{szegedy2014intriguing,goodfellow2015explaining,papernot2016limitations,moosavi2016deepfool,carlini2017towards,madry2018towards,huang2017adversarial,kos2017delving,lin2017tactics,pattanaik2018robust,gleave2020adversarial,liang2020efficient,wei2018transferable,liang2021generate,liang2022parallel,liang2022large,liu2023x,wang2022bet,wang2025your}. Stronger variants have since been developed from several angles, including robust state-observation formulations~\citep{zhang2020robust,zhang2021robust}, real-time and universal observation attacks~\citep{tekgul2022realtime,moosavi2017uap,wang2023diversifying,liu2023improving}, policy-aware attackers~\citep{sun2022strongest}, detectability-aware stealthy attacks~\citep{franzmeyer2024illusory}, and policy-distribution-based methods~\citep{dapgd2025,korkmaz2023adversarial,bai2025rat}. However, most of these methods still optimize perturbations directly in pixel space and treat each frame as an essentially independent target. That assumption is mismatched with visual world-model agents: what matters is not only whether one frame is perturbed, but whether the perturbation keeps steering the recurrent latent state under a realistic per-frame compute budget.

We study this setting in a white-box, causal, online attack regime against a frozen DreamerV3 victim. The attacker may modify only the current image observation and must produce the perturbation before the next environment step. Across DMC walker walk (DeepMind Control Suite~\citep{tassa2018dmc}), Atari Pong, and Crafter, we observe two consistent phenomena. First, naive per-frame full-pixel projected gradient descent (PGD)~\citep{madry2018towards} is a weak primitive here: recurrent latent dynamics dilute isolated high-frequency perturbations, so greedy frame-wise attacks struggle to accumulate influence over long horizons. Second, effective attack directions do not spread uniformly across the full pixel space. When we stack critic-guided perturbations across time, their energy concentrates in a surprisingly low-dimensional subspace. This echoes subspace and low-frequency findings from supervised learning~\citep{yan2019subspace,guo2019low,moosavi2017uap,lou2024hide,jing2024deep}, but the structure here is induced by the victim's own value landscape rather than by input statistics.

These observations suggest a different attack strategy: identify value-sensitive directions once, then reuse them causally over time. We instantiate this idea as \textbf{CIVA}. CIVA first collects critic-guided perturbations offline and applies singular value decomposition (SVD) to extract a low-dimensional, victim-aware attack subspace. At deployment time, it no longer searches the full pixel space at every frame; instead, it optimizes only the subspace coefficients and smooths them with an exponential moving average before lifting them back to pixels. This directly addresses the core difficulty of attacking world-model agents under a tight $\ell_\infty$ budget: the perturbation must remain strong enough to influence recurrent latent dynamics while also staying computationally cheap, temporally coherent, and visually subtle~\citep{he2023sa,zhang2024visual,wang2025black,kong2024patch,zhao2021multi}. Across continuous control, discrete control, and open-ended tasks, CIVA yields stronger and more stable degradation than representative observation-space baselines~\citep{zhang2020robust,tekgul2022realtime,sun2022strongest,franzmeyer2024illusory,dapgd2025}, while preserving favorable temporal smoothness and perceptual similarity. Ablations further show that the gains come from the combination of a critic-induced subspace, online optimization constrained to that subspace, and temporal smoothing on its coefficients.

Our contributions are threefold:
\begin{itemize}
\item We identify a key mismatch between conventional frame-wise observation attacks and visual world-model agents: because decisions depend on recurrent latent state, successful attacks must shape the victim's internal representation over time rather than only degrade individual frames~\citep{liu2024OPT,liu2023contrastive}.
\item We propose CIVA, a critic-induced value-subspace attack that extracts a low-rank victim-aware perturbation basis offline and performs efficient causal online optimization with temporal smoothing inside that subspace.
\item We evaluate CIVA on DMC walker walk, Atari Pong, and Crafter across attack effectiveness, action-distribution shift, temporal variation, and perceptual similarity, showing that low-dimensional critic-aligned perturbations are more effective for world-model agents than direct full-pixel frame-wise optimization.
\end{itemize}

%% file: Sec/2-related-work.tex
\paragraph{World-model reinforcement learning.}
World models build policies around a recurrent latent dynamics model that summarizes past observations and supports multi-step prediction~\citep{ha2018worldmodels}. The Dreamer line demonstrates that this recipe scales across continuous control, Atari, and open-ended visual domains under a largely unified configuration~\citep{hafner2019dreamerv1,hafner2020dreamerv2,hafner2023mastering}. Parallel efforts extend the same general direction to planning with learned models~\citep{schrittwieser2020muzero}, transformer-~\citep{micheli2023iris,robine2023twm,zhang2023storm} and diffusion-based world models~\citep{alonso2024diamond}, large-scale generative environments~\citep{bruce2024genie}, scalable model-based control~\citep{hansen2024tdmpc2}, and robotic deployment~\citep{wu2023daydreamer}; recent surveys summarize this broader trend~\citep{ding2024worldmodelsurvey,guo2026world,xu2026ctrlattack}. For our purposes, the important point is not only that these agents are powerful, but that their decision process is mediated by a recurrent latent state. That makes their adversarial surface fundamentally temporal rather than frame-local.

\paragraph{Adversarial attacks on deep policies.}
Adversarial vulnerability in supervised learning~\citep{szegedy2014intriguing,goodfellow2015explaining,papernot2016limitations,moosavi2016deepfool,carlini2017towards,madry2018towards} carries over directly to RL, where small perturbations in observations can cause severe behavioral degradation~\citep{huang2017adversarial,kos2017delving,lin2017tactics,pattanaik2018robust}. Follow-up work has strengthened this line through robust state-observation formulations~\citep{zhang2020robust,zhang2021robust}, real-time and universal perturbations~\citep{tekgul2022realtime,wang2023diversifying,liu2023improving}, policy-aware attackers~\citep{sun2022strongest}, detectability-aware attacks~\citep{franzmeyer2024illusory,he2023sa,kong2024patch}, and policy-distribution-based objectives~\citep{dapgd2025,zhang2024visual,wang2025black}. Other work studies adversarial agents or robustness notions more broadly in multi-agent and deep-RL settings~\citep{gleave2020adversarial,korkmaz2023adversarial,bai2025rat,liu2024OPT,liu2023contrastive}. These methods provide strong attack baselines, but most still optimize directly in pixel space and effectively target each frame separately. Our setting differs because the victim is a visual world-model agent whose behavior depends on how perturbations accumulate inside recurrent latent dynamics.

\paragraph{Robust RL and defenses.}
From the defense side, robust RL has been studied via adversarial training with a learned attacker~\citep{pinto2017robust,zhang2021robust}, regularization against worst-case observation perturbations~\citep{oikarinen2021radial}, worst-case-aware training without explicit adversarial rollouts~\citep{liang2022wocar}, and game-theoretic formulations for temporally coupled disturbances~\citep{liang2024gametheoretic}. This literature mainly asks how to train agents that remain reliable under perturbation. Our question is different: before defending visual world-model agents, we need to understand what strong causal observation attacks look like against them in the first place, and whether their recurrent latent structure induces a different geometry of effective perturbations.

\paragraph{Structured, low-dimensional, and temporally coherent attacks.}
Another relevant thread asks whether adversarial perturbations really need the full pixel space. In supervised learning, universal perturbations~\citep{moosavi2017uap}, low-frequency attacks~\citep{guo2019low}, and subspace attacks~\citep{yan2019subspace,lou2024hide,jing2024deep} all show that effective directions often lie in much lower-dimensional or smoother spaces than raw pixels. In video and streaming settings, temporally coordinated attacks can be more effective and less perceptible than independent frame-wise perturbations~\citep{wei2019sparse,li2019stealthy}. RL work such as real-time universal perturbations and policy-aware attackers similarly introduces cross-frame structure to improve efficiency or attack strength~\citep{tekgul2022realtime,sun2022strongest,dapgd2025}. CIVA is related in spirit but differs in the origin of the subspace and in how it is deployed. Our attack basis is not hand-crafted, random, or derived from observation statistics; it is extracted directly from the victim critic's attack gradients collected on rollouts of the target agent. The resulting basis is then reused causally across frames, with temporal coherence enforced by smoothing the subspace coefficients. This ties low-dimensional attack structure directly to the value landscape of a recurrent world-model agent.

%% file: Sec/3-threat-model.tex
\subsection{Victim Model}
We attack visual world-model reinforcement learning agents.
The victim is a pretrained \textbf{DreamerV3} policy~\citep{hafner2023mastering}.
At each step the agent receives an RGB image
$o_t \in [0,255]^{H\times W\times C}$ with $H{=}W{=}64$ and $C{=}3$.
The world model updates a recurrent latent state from this image.
The actor head then outputs an action, and the critic head outputs a
categorical value distribution.
We use the same victim setup on three tasks that cover different
decision regimes:
DMC walker walk (continuous control),
Atari Pong (discrete control),
and Crafter (open-ended world).
The victim weights and the environment dynamics are frozen during the
attack. The attacker does not retrain or fine-tune the victim.

\subsection{Attacker's Capability}
\label{sec:capability}
The attacker can only modify the image observation that is fed into the
agent. It cannot touch the actions, rewards, environment state, or the
recurrent latent.
We use a \emph{white-box online} setting.
The attacker has full access to the policy and value heads and can
backpropagate through them.
The attacker is also \emph{causal}: at frame $t$ it can only use the
current and past observations, and it must produce $\delta_t$ before the
next environment step.
To match the comparison protocol in Section~\ref{sec:experiment}, this
causal constraint applies at \emph{deployment time} to all methods,
including baselines. Methods that require one-time preprocessing (e.g.,
universal perturbation fitting or subspace extraction) are allowed to use
clean rollouts from the same frozen victim before deployment, but the
resulting parameters (e.g., the value-subspace basis $V_r$ used by CIVA)
are then frozen for the entire evaluation; no method may access future
test observations, alter the environment.
Each perturbation is bounded by an $\ell_\infty$ pixel budget,
$\lVert \delta_t \rVert_\infty \le \varepsilon$.
We use $\varepsilon = 24/255$ throughout this paper.
A larger budget makes the perturbation visually obvious, while a much
smaller budget barely degrades the agent.
We find $24/255$ to be a good trade-off between attack strength and
visual stealth.
The attacker also has only a small per-frame compute budget,
i.e.\ a few gradient steps per observation.
A practical attack must therefore be both parameter-efficient and
step-efficient.

\subsection{Attacker's Goal}
Let $T$ be the episode horizon and $t_{\text{start}}$ the attack onset.
The attacker minimizes the cumulative episodic return:
\begin{equation}
\min_{\{\delta_t\}_{t=t_{\text{start}}}^{T}}\;
   \sum_{t=t_{\text{start}}}^{T} r_t
\quad \text{s.t.}\quad
\lVert \delta_t \rVert_\infty \le \varepsilon,\ \
o_t + \delta_t \in [0,255]^{H\times W\times C}.
\end{equation}
The reward is not differentiable through the environment.
We therefore optimize victim-exposed differentiable surrogates, such as
value-based, action-based, or reward-head objectives. Baselines are
evaluated under the same threat model and keep their method-specific
optimization forms when applicable, but all optimization signals are
restricted to victim-exposed differentiable quantities and never use
true environment reward gradients. Beyond reducing return, an effective
attack should also be \emph{stealthy}: temporally coherent, spatially
structured, and visually close to the clean observation.

\subsection{Challenges}
\label{sec:challenges}
The above setting raises three coupled challenges that motivate our
design.
\textbf{(C1) Recurrent dilution.}
A world-model agent integrates observations through a recurrent latent
state. This latent state smooths out isolated high-frequency pixel
noise. As a result, single-frame greedy attacks struggle to accumulate
effect over a long horizon.
\textbf{(C2) Cost of full-pixel per-frame PGD.}
Running multi-step white-box PGD in the full
$D = H\!\times\!W\!\times\!3$ pixel space at every frame is expensive.
It also produces unstructured and temporally jittery perturbations.
This is incompatible with the small per-frame compute budget stated in
Section~\ref{sec:capability}.
\textbf{(C3) Budget--effectiveness--stealth trilemma.}
Under a tight $\ell_\infty$ budget, we want strong return suppression,
temporal smoothness, and visual naturalness at the same time.
Naive PGD usually sacrifices at least one of the three.
These three challenges point to the same need:
a low-dimensional, data-driven perturbation parameterization that is
shared across frames and explicitly enforces temporal coherence.
We develop such a design in Section~\ref{sec:method}.

%% file: Sec/4-method.tex
\begin{figure}[t]
  \centering
   \includegraphics[width=0.92\linewidth]{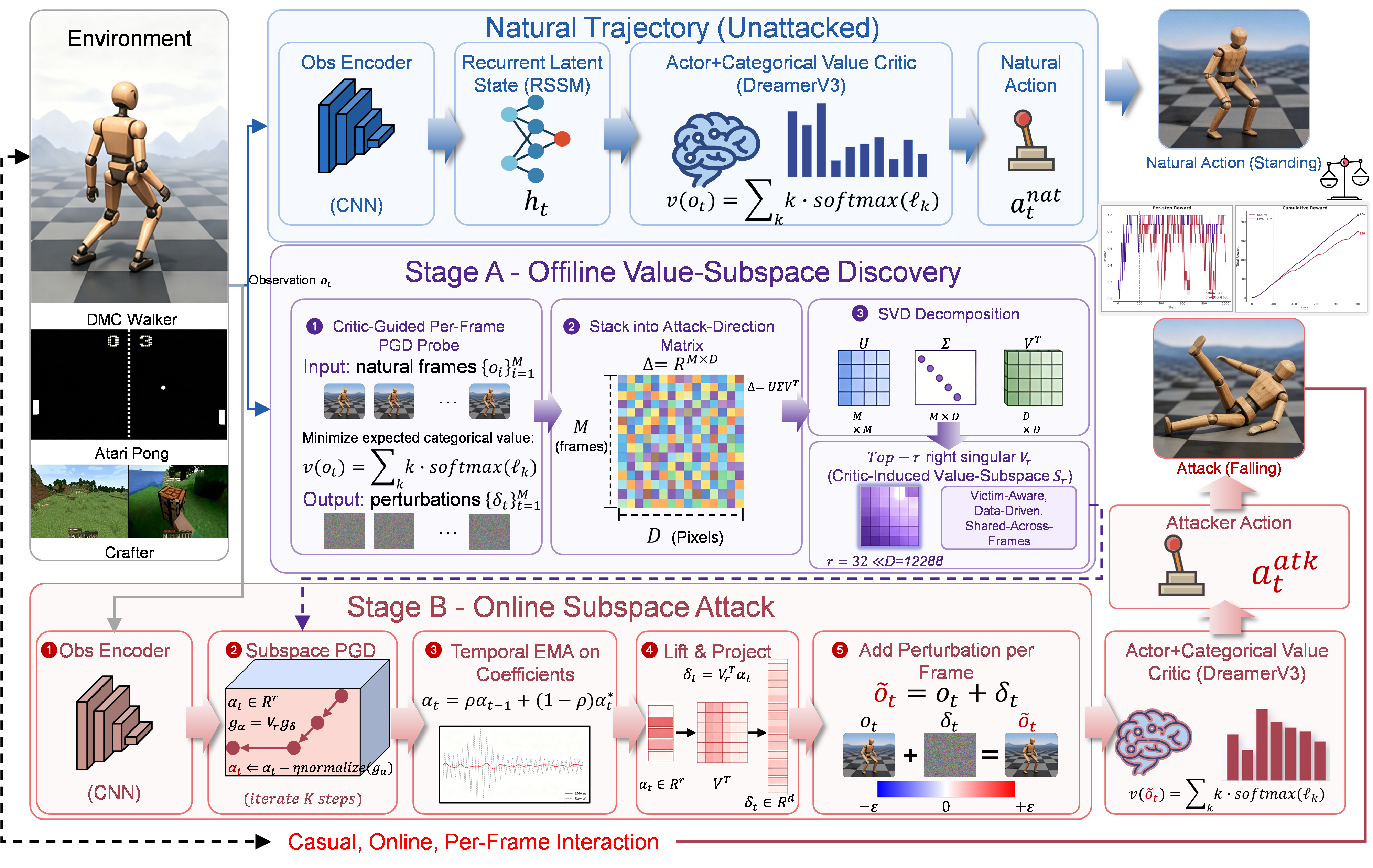}
   \caption{\textbf{CIVA framework.} Stage~A extracts a critic-induced value-subspace from offline PGD probes; Stage~B performs online subspace-PGD with EMA smoothing.}
  \label{fig:civa}
\end{figure}

Motivated by the challenges in Section~\ref{sec:challenges}, we propose
\textbf{CIVA} (Critic-Induced Value-Subspace Attacks).
Unlike prior subspace attacks that build the search space from input
statistics (e.g., PCA over observations or hand-crafted DCT bases)~\citep{yan2019subspace,guo2019low},
CIVA constructs its low-dimensional subspace directly from the
victim critic's own attack gradients.
As shown in Figure~\ref{fig:civa}, CIVA proceeds in two stages.
\textbf{Stage~A (offline)} probes the victim with critic-guided
per-frame PGD on a clean rollout, collects the resulting pixel
perturbations, and extracts a low-dimensional value-subspace via SVD.
\textbf{Stage~B (online)} performs per-frame PGD only inside this
subspace, and uses an exponential moving average (EMA) on the
subspace coefficients to enforce temporal coherence. We detail the
two stages in Section~\ref{sec:civa-offline} and
Section~\ref{sec:civa-online}.

\subsection{Offline Value-Subspace Discovery}
\label{sec:civa-offline}

\paragraph{Critic-guided per-frame PGD probing.}
The strong baseline implied by Section~\ref{sec:capability} is
per-frame full-pixel PGD: at every frame the attacker searches the
full $D = H\!\times\!W\!\times\!C$ pixel space with multiple gradient
steps~\citep{madry2018towards,tekgul2022realtime}. The resulting perturbations are independent across frames,
temporally jittery, and sit in a $D = 12288$ dimensional space.
A natural question is whether all $D$ pixel directions are truly
needed, or whether the effective attack directions concentrate in a
much lower-dimensional subspace.
To answer this, we first \emph{collect} a set of critic-guided pixel
perturbations on a clean rollout. We start the collection only after a short
$200$-step warm-up so that the recurrent latent state has stabilised.

The DreamerV3 critic outputs a categorical value distribution
$p(k\mid o_t)$ over $K$ symlog-spaced bins~\citep{hafner2023mastering}. We do \emph{not} need to
decode this distribution back into a calibrated value: any quantity
that is monotone in the agent's predicted return suffices as an attack
signal. We therefore use the expected \emph{bin index}
\begin{equation}
v(o_t) \;=\; \sum_{k=1}^{K} k \cdot
   \mathrm{softmax}(\ell_k(o_t)),
\label{eq:value-loss}
\end{equation}
as a cheap differentiable surrogate (since the bins are sorted by
value, lowering $v(o_t)$ shifts probability mass toward lower-value
bins), and run $L$ steps of $\ell_\infty$ PGD per frame:
\begin{equation}
\delta^{(\ell+1)} \;=\; \Pi_{\mathcal{B}_\varepsilon}\!\Big(
   \delta^{(\ell)} - \eta \cdot \mathrm{normalize}\big(
   \nabla_\delta v(o_t + \delta^{(\ell)})\big)\Big),
\label{eq:pgd-collect}
\end{equation}
where $\mathrm{normalize}(g) := g / \sqrt{\mathbb{E}[g^2]}$ is
RMS-normalization of the (sub)gradient~\citep{tieleman2012rmsprop} and
$\mathcal{B}_\varepsilon = \{\delta : \lVert\delta\rVert_\infty
\le \varepsilon\}$ is composed with the pixel-range projection
$o_t + \delta \in [0,255]^{H\times W\times C}$. We denote the final
perturbation as $\delta_i := \delta^{(L)}$ and repeat this for
$M$ frames. Stacking the flattened perturbations gives the
\emph{attack-direction matrix}
\begin{equation}
\Delta \;=\; [\delta_1, \dots, \delta_M]^{\top} \in \mathbb{R}^{M\times D}.
\label{eq:delta-matrix}
\end{equation}
Each row of $\Delta$ is, by construction, a pixel direction that
\emph{empirically} drives the victim critic toward lower predicted
value. It is neither an observation, nor a hand-crafted low-frequency
prior, but the victim's own ``where to push pixels'' signal.

\paragraph{Constructing the critic-induced value-subspace.}
Given $\Delta$, we take a truncated SVD~\citep{eckart1936approximation}
\begin{equation}
\Delta \;=\; U\Sigma V^{\top},
\qquad
V_r \;:=\; V[:,{:}r] \in \mathbb{R}^{D\times r}.
\label{eq:svd}
\end{equation}
The columns of $V_r$ are the top-$r$ right singular vectors of $\Delta$ and
form an orthonormal basis of an $r$-dimensional subspace of $\mathbb{R}^D$;
we call its column span
$\mathcal{S}_r := \mathrm{col}(V_r)$ the
\emph{Critic-Induced Value-Subspace}. CIVA confines all online
perturbations to $\mathcal{S}_r$. The basis $V_r$ is computed once offline
and remains frozen at evaluation time. With $r = 32 \ll D = 12288$, sharing this
victim-aware, spatially structured basis across frames directly addresses challenges:
it shrinks the per-frame search space, avoids the high-frequency artifacts of
independent pixel-level PGD, and keeps consecutive perturbations on a common
direction set that the EMA in Section~\ref{sec:civa-online} can smooth without
losing critic-effectiveness.

\paragraph{Relation to existing subspace attacks.}
Prior subspace attacks typically derive their search spaces from input statistics, such as PCA or hand-crafted DCT bases, or from random/query-saving projections in black-box settings. CIVA instead extracts its subspace directly from the victim critic's gradients, making $V_r$ victim-aware by construction. This distinction matters in our online recurrent setting, where temporal coherence is central and is evaluated explicitly in Section~\ref{sec:experiment} through comparison with a random-basis ablation.

\subsection{Online Subspace Attack}
\label{sec:civa-online}

Given $V_r$, we now describe how CIVA produces the per-frame
perturbation $\delta_t$ during deployment.

\paragraph{Value-based attack objective.}
The true reward $r_t$ is non-differentiable through the environment
(see Section~\ref{sec:capability}), so we minimize the same
critic-based surrogate used in Stage~A. Given the current observation
$o_t$ and a perturbation $\delta_t$, we define the per-frame loss
\begin{equation}
\mathcal{L}_{\mathrm{val}}(\delta_t; o_t)
\;=\; v(o_t + \delta_t),
\label{eq:l-val}
\end{equation}
where $v(\cdot)$ is the same expected value as in
Eq.~\eqref{eq:value-loss}. Minimizing $\mathcal{L}_{\mathrm{val}}$
makes the agent believe its future return is lower than it actually
is, and indirectly induces suboptimal actions. Using the same
surrogate in Stage~A and Stage~B keeps the directions in $V_r$
semantically aligned with the online optimization target.

\paragraph{Subspace reparameterization.}
We reparameterize the per-frame perturbation as a low-dimensional
coefficient~\citep{ilyas2018blackbox} $\alpha_t \in \mathbb{R}^r$:
\begin{equation}
\delta_t \;=\; V_r\, \alpha_t.
\label{eq:reparam}
\end{equation}
Online PGD then operates in $\mathbb{R}^r$ rather than $\mathbb{R}^D$.
At each step we first take a pixel-space gradient
$g_\delta = \nabla_\delta \mathcal{L}_{\mathrm{val}}$, project it
into the subspace, and apply a normalized gradient step:
\begin{equation}
g_\alpha \;=\; V_r^{\top}\, g_\delta,
\qquad
\alpha_t^{(\ell+1)} \;=\; \alpha_t^{(\ell)} - \eta \cdot
   \mathrm{normalize}(g_\alpha).
\label{eq:subspace-pgd}
\end{equation}
The $\ell_\infty$ budget and the pixel-range constraint are still
enforced in pixel space: after each update we lift $\alpha_t$ to
$\delta_t$, project onto $\mathcal{B}_\varepsilon$ and onto
$[0,255]^{H\times W\times C}$, and re-project back to $\alpha_t$.
This approximately enforces the subspace constraint, the $\ell_\infty$
budget, and pixel validity at the same time; in practice the residual
reprojection error is below $10^{-3}$ in pixel scale and does not
affect the reported results.

\paragraph{Temporal coherence via subspace EMA.}
C3 in Section~\ref{sec:capability} requires the perturbation
sequence to be temporally coherent rather than just frame-wise
strong. We enforce this directly on the subspace coefficients via an
exponential moving average~\citep{polyak1992acceleration}:
\begin{equation}
\alpha_t \;=\; \rho \cdot \alpha_{t-1}
   \;+\; (1-\rho) \cdot \alpha_t^{\star},
\label{eq:ema}
\end{equation}
where $\alpha_t^{\star}$ is the per-frame PGD optimum from
Eq.~\eqref{eq:subspace-pgd}, and $\rho \in [0,1]$ controls the
smoothing strength. With $\rho \to 1$ the perturbation barely
changes across frames (very smooth, slightly weaker); with $\rho \to 0$
CIVA reduces to independent per-frame attack (strongest, but most
jittery).
We stress that the EMA is meaningful precisely \emph{because} it is
applied in the subspace. A pixel-space EMA on $\delta_t$ would act as
a temporal low-pass filter and could smooth the perturbation off the
critic-effective direction set. Inside $\mathcal{S}_r$, every
$\alpha_t$ is by construction a critic-effective direction, so
temporal smoothing does not trade away attack strength. This is the
direct payoff of sharing $V_r$ across frames, as anticipated in
Section~\ref{sec:civa-offline}.

\paragraph{Joint online procedure.}
Putting Eq.~\eqref{eq:l-val}--\eqref{eq:ema} together, CIVA produces
$\delta_t$ at frame $t$ as follows.
(i) Warm-start from the previous (post-EMA) coefficient $\alpha_{t-1}$
and run a small number of subspace-PGD steps on
$\mathcal{L}_{\mathrm{val}}$ (Eq.~\eqref{eq:subspace-pgd}) to obtain
$\alpha_t^{\star}$.
(ii) Smooth via Eq.~\eqref{eq:ema} to get $\alpha_t$.
(iii) Lift to $\delta_t = V_r\,\alpha_t$ and project onto the
$\ell_\infty$ ball and the pixel range.
(iv) Feed $o_t + \delta_t$ to the victim and step the environment.
The procedure is causal, online, parameter-efficient (only $r$
scalars per frame), and step-efficient (a few PGD steps per frame),
satisfying every constraint of Section~\ref{sec:challenges}.

%% file: Sec/5-experiment.tex
\label{sec:experiment}

\subsection{Experimental Setup}
\label{sec:exp-setup}

\paragraph{Environments and victim.}
We evaluate on three visual control benchmarks covering distinct action regimes: DMC  walker walk~\citep{tassa2018dmc} (continuous), Atari Pong~\citep{bellemare2013ale,mnih2015human} (discrete), and Crafter~\citep{hafner2021crafter} (open-ended survival). All environments deliver $64\!\times\!64\!\times\!3$ RGB observations ($D{=}12{,}288$) with a 1{,}000-step episode horizon. The victim is DreamerV3~\citep{hafner2023mastering} trained per task with the official configuration; the resulting checkpoint is frozen for all attacks.

\paragraph{Baselines and attack budget.}
We compare CIVA against five representative observation-space attackers: MAD~\citep{zhang2020robust}, UAP-RL~\citep{tekgul2022realtime}, PA-AD~\citep{sun2022strongest}, Illusory~\citep{franzmeyer2024illusory}, and DAPGD~\citep{dapgd2025}. All methods share an $\ell_\infty$ per-frame budget $\varepsilon{=}24/255$ and identical white-box access to the frozen victim.

\paragraph{CIVA hyperparameters.}
\textbf{Stage~A} (offline) collects $M{=}600$ critic-guided per-frame PGD perturbations with $L{=}8$ inner steps and step size $\eta{=}\varepsilon/4$, after a 200-step warm-up that lets the recurrent latent state stabilise. SVD on $\Delta\in\mathbb{R}^{600\times 12288}$ yields the top $r{=}32$ right singular vectors that form the columns of $V_r\in\mathbb{R}^{12288\times 32}$. \textbf{Stage~B} (online) reuses $L{=}8$ and $\eta{=}\varepsilon/4$ but optimises only the coefficient vector $\alpha_t\in\mathbb{R}^{r}$, smoothed by an EMA with momentum $\rho{=}0.75$ before being lifted back to pixels. The basis $V_r$ is computed once and frozen at evaluation time.

\paragraph{Evaluation protocol.}
Each method is evaluated over $5$ independent random seeds (one episode per seed) and we report the across-seed means of episode reward, the reward drop ratio (\textbf{Drop\%}) relative to the natural policy following the convention of~\citep{pattanaik2018robust,zhang2020robust}, the Kullback--Leibler divergence~\citep{kullback1951information} between clean and attacked action distributions (\textbf{Action-KL}) as used in policy-distribution attacks~\citep{sun2022strongest,dapgd2025}, the mean absolute frame-to-frame change of the perturbation (\textbf{TempAbs}) inspired by temporal-coherence measures for video adversarial attacks~\citep{wei2019sparse,li2019stealthy}, and SSIM~\citep{wang2004ssim} between clean and perturbed observations. All experiments run on a single NVIDIA RTX PRO 6000 Blackwell GPU (96\,GB).

\subsection{Main Results}
\label{sec:exp-main}

Tables~\ref{tab:dmc} and~\ref{tab:pong} report main attack results on DMC walker walk (continuous control) and Atari Pong (discrete control). Crafter results are deferred to Appendix~\ref{app:crafter}.

\input{Table/tab2.tex}
\input{Table/tab3.tex}

\begin{figure}[t]
  \centering
  \includegraphics[width=\linewidth]{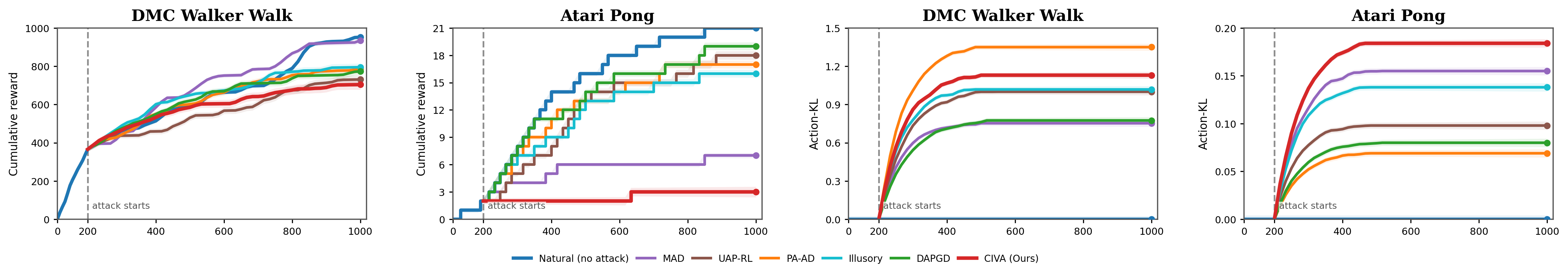}
  \caption{\textbf{Attack dynamics over time.} Cumulative reward (left pair) and Action-KL (right pair) on DMC walker walk and Atari Pong. The dashed line marks the attack start at $t{=}200$.}
  \label{fig:reward-curves}
\end{figure}

\paragraph{Attack effectiveness.}
Under the same $\ell_\infty{=}24/255$ budget, Tables~\ref{tab:dmc} and~\ref{tab:pong} show that \textbf{CIVA achieves the strongest attack effectiveness on both tasks}. On DMC walker walk, it attains the lowest attacked reward ($703.97$) and the highest Drop\% ($26.07\%$), outperforming the strongest baseline UAP-RL ($23.25\%$). On Atari Pong, the margin is larger: CIVA reaches $85.71\%$ Drop\%, whereas the runner-up MAD reaches $66.67\%$ and all other baselines remain below $25\%$. Figure~\ref{fig:reward-curves} is consistent with the table results: after the attack starts at $t{=}200$, CIVA yields the slowest cumulative reward growth on both tasks, finishing lowest on DMC and plateauing near reward $3$ on Pong. Overall, CIVA not only produces the best final reward suppression, but also degrades performance more persistently over time.

\paragraph{Behavioural deviation.}
The Action-KL column and the right half of Figure~\ref{fig:reward-curves} show that CIVA produces the strongest behavioural deviation, reaching $1.130$ on DMC and $\mathbf{0.184}$ on Pong. DMC also shows that KL alone is insufficient: PA-AD attains the largest Action-KL ($1.35$) but achieves only a $17.38\%$ reward drop. By constraining perturbations to the critic-relevant subspace, CIVA induces policy shifts that are better aligned with long-horizon value degradation.

\paragraph{Temporal coherence and visual stealth.}
Per-frame baselines pay a heavy temporal cost: PA-AD, Illusory and DAPGD report TempAbs of $10$--$20$ on DMC and $4$--$17$ on Pong, one to two orders of magnitude above CIVA ($0.646$ and $0.394$). UAP-RL has lower TempAbs only because it injects a static pattern, which caps its Pong drop at $14.29\%$. \textbf{CIVA delivers the strongest reward drop while maintaining very low temporal variation}, improving TempAbs by roughly $10\times$ over per-frame baselines while keeping SSIM within $0.04$--$0.30$ of the most stealth-oriented baseline (MAD). On the same GPU, the online subspace optimization further reduces per-frame attack latency by roughly an order of magnitude relative to full-pixel per-frame PGD (V2 in Section~\ref{sec:exp-ablation}), since the inner PGD operates over $r{=}32$ scalars instead of $D{=}12{,}288$ pixels.

\FloatBarrier

\subsection{Analysis of Policy Decision Shifts}
\label{sec:exp-decision-shifts}

\begin{figure}[htbp]
  \centering
  \includegraphics[width=\linewidth]{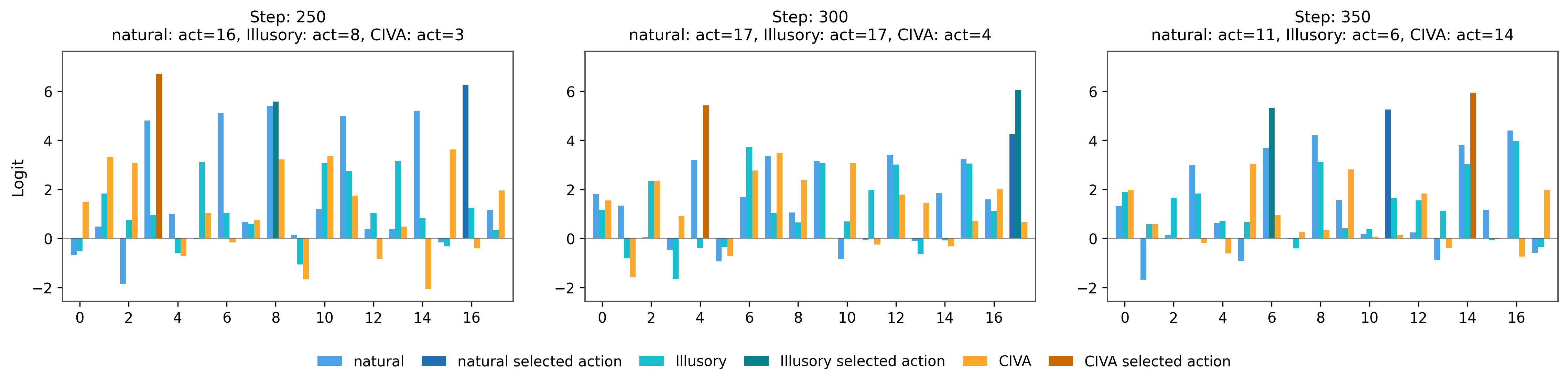}
  \caption{\textbf{Policy-logit shifts across attacks on Atari Pong.} Dark bars mark the selected actions.}
  \label{fig:policy-logit-shifts}
\end{figure}

To complement the aggregate reward and Action-KL metrics, Figure~\ref{fig:policy-logit-shifts} compares the victim's action logits under the natural policy, Illusory, and CIVA at three representative attack-time steps. Illusory changes the logit landscape but can leave the selected action unchanged, as at step~300. In contrast, CIVA consistently makes a different action dominant in these examples, showing that the critic-induced subspace translates distributional shifts into concrete decision changes. This also explains why CIVA's large Action-KL in Tables~\ref{tab:dmc} and~\ref{tab:pong} is coupled with larger reward degradation: the shifted distributions cross the policy's action-selection boundary at decision-critical frames.

\FloatBarrier

\subsection{Ablation Study}
\label{sec:exp-ablation}

We ablate each component of CIVA on DMC walker walk; Tables~\ref{tab:ablation-reward} and~\ref{tab:ablation-stealth} report effectiveness and stealth. From the full model (V0), each variant changes one component: V1 replaces the critic-induced subspace with a random orthonormal basis of the same rank; V2 removes the subspace and uses full-pixel PGD; V3 removes the EMA ($\rho{=}0$); V4 replaces the critic-value surrogate with the victim's reward-prediction-head gradient.

\input{Table/tab4.tex}

\input{Table/tab5.tex}

\paragraph{Critic-induced subspace.}
A random basis of the same rank (V1) drops effectiveness from $26.07\%$ to $6.90\%$ and KL from $1.130$ to $0.058$. Full-pixel critic-PGD (V2) is also weak ($12.61\%$ drop) and inflates TempAbs to $3.235$, the largest in the table. \textbf{Low rank alone is not enough, and critic guidance alone is not enough either}; the gain comes from aligning a low-rank basis with critic-sensitive directions.

\paragraph{Temporal EMA.}
Disabling the EMA (V3) gives a marginal gain in effectiveness ($+0.77$ pp) but \textbf{nearly triples TempAbs} ($1.7496$ vs.\ $0.646$) and lowers SSIM by $3.5$ points. Since CIVA targets the joint optimum of effectiveness \emph{and} stealth, we keep the EMA in V0: the small reward-drop loss is outweighed by the much smoother and more imperceptible perturbation sequence.

\paragraph{Critic value vs.\ reward head.}
Replacing the critic-value gradient with the reward-head gradient (V4) drops effectiveness to $4.84\%$, the weakest variant, and Action-KL also falls below V0 ($1.054$ vs.\ $1.130$). Although V4's stealth indicators (TempAbs $0.542$, SSIM $0.833$) look favourable, this is precisely because the reward-head signal points along directions that move the world model's one-step predicted reward without affecting the long-horizon return that drives the policy: the perturbations are easy to make smooth and visually subtle, but no longer functionally adversarial. Taking effectiveness, behavioural deviation, and stealth together, V0 (full CIVA) gives the best overall trade-off among all variants.

%% file: Table/tab2.tex
\begin{table}[t]
  \centering
  \caption{\textbf{Attack results on DMC walker walk.} ``Natural'' is the unattacked baseline; arrows show the favorable direction for the attacker. \textbf{Bold} = best, \underline{underline} = second best.}
  \label{tab:dmc}
  \small
  \setlength{\tabcolsep}{4pt}
  \renewcommand{\arraystretch}{1.15}
  \arrayrulecolor{civarule}
  \resizebox{\linewidth}{!}{%
  \begin{tabular}{llccccc}
    \toprule
    \rowcolor{civablue}
    \color{white}\textbf{Method} & \color{white}\textbf{Venue / Year} & \color{white}\textbf{Episode Reward $\downarrow$} & \color{white}\textbf{Drop\% $\uparrow$} & \color{white}\textbf{Action-KL $\uparrow$} & \color{white}\textbf{TempAbs $\downarrow$} & \color{white}\textbf{SSIM $\uparrow$} \\
    \midrule
    Natural (no attack) & ---            & 952.25 & ---     & ---    & ---    & 1.000 \\
    \midrule
    \rowcolor{civastripe} MAD~\citep{zhang2020robust}                 & NeurIPS 2020   & 934.71 & 1.84\%  & 0.753  & 9.126  & \textbf{0.748} \\
    UAP-RL~\citep{tekgul2022realtime}              & ESORICS 2022   & \underline{730.86} & \underline{23.25\%} & 1.000  & \textbf{0.00} & 0.656 \\
    \rowcolor{civastripe} PA-AD~\citep{sun2022strongest}               & ICLR 2022      & 786.75 & 17.38\% & \textbf{1.35}   & 19.97  & 0.35  \\
    Illusory~\citep{franzmeyer2024illusory}            & ICLR 2024      & 795.13 & 16.50\% & 1.018  & 10.97  & 0.445 \\
    \rowcolor{civastripe} DAPGD~\citep{dapgd2025}               & ICASSP 2025    & 773.33 & 18.79\% & 0.775  & 12.31  & 0.501 \\
    \midrule
    \rowcolor{civarow} \textbf{CIVA (Ours)} & \textbf{2026} & \textbf{703.97} & \textbf{26.07\%} & \underline{1.130} & \underline{0.646} & \underline{0.710} \\
    \bottomrule
  \end{tabular}%
  }
  \arrayrulecolor{black}
\end{table}

%% file: Table/tab3.tex
\begin{table}[t]
  \centering
  \caption{\textbf{Attack results on Atari Pong.}}
  \label{tab:pong}
  \small
  \setlength{\tabcolsep}{4pt}
  \renewcommand{\arraystretch}{1.15}
  \arrayrulecolor{civarule}
  \resizebox{\linewidth}{!}{%
  \begin{tabular}{llccccc}
    \toprule
    \rowcolor{civablue}
    \color{white}\textbf{Method} & \color{white}\textbf{Venue / Year} & \color{white}\textbf{Episode Reward $\downarrow$} & \color{white}\textbf{Drop\% $\uparrow$} & \color{white}\textbf{Action-KL $\uparrow$} & \color{white}\textbf{TempAbs $\downarrow$} & \color{white}\textbf{SSIM $\uparrow$} \\
    \midrule
    Natural (no attack) & ---            & 21.0 & ---     & ---   & ---   & 1.000 \\
    \midrule
    \rowcolor{civastripe} MAD~\citep{zhang2020robust}                 & NeurIPS 2020   & \underline{7.0}  & \underline{66.67\%} & \underline{0.155} & 4.357 & \textbf{0.864} \\
    UAP-RL~\citep{tekgul2022realtime}              & ESORICS 2022   & 18.0 & 14.29\% & 0.098 & \textbf{0.000} & 0.344 \\
    \rowcolor{civastripe} PA-AD~\citep{sun2022strongest}               & ICLR 2022      & 17.0 & 19.05\% & 0.069 & 4.029 & 0.697 \\
    Illusory~\citep{franzmeyer2024illusory}            & ICLR 2024      & 16.0 & 23.81\% & 0.138 & 9.570 & 0.599 \\
    \rowcolor{civastripe} DAPGD~\citep{dapgd2025}               & ICASSP 2025    & 19.0 & 9.52\%  & 0.080 & 17.194 & \underline{0.719} \\
    \midrule
    \rowcolor{civarow} \textbf{CIVA (Ours)} & \textbf{2026} & \textbf{3.0} & \textbf{85.71\%} & \textbf{0.184} & \underline{0.394} & 0.566 \\
    \bottomrule
  \end{tabular}%
  }
  \arrayrulecolor{black}
\end{table}

%% file: Table/tab4.tex
\begin{table}[t]
  \centering
  \caption{\textbf{Ablation: reward drop on DMC walker walk.} The \textbf{Online} and \textbf{EMA} columns indicate whether subspace-restricted online PGD and temporal EMA are enabled.}
  \label{tab:ablation-reward}
  \small
  \setlength{\tabcolsep}{4pt}
  \renewcommand{\arraystretch}{1.15}
  \arrayrulecolor{civarule}
  \resizebox{\linewidth}{!}{%
  \begin{tabular}{llllccrr}
    \toprule
    \rowcolor{civablue}
    \color{white}\textbf{ID} &
    \color{white}\textbf{Variant} &
    \color{white}\textbf{(A) Signal} &
    \color{white}\textbf{(B) Subspace} &
    \color{white}\textbf{Online} &
    \color{white}\textbf{EMA} &
    \color{white}\textbf{EpReward $\downarrow$} &
    \color{white}\textbf{Drop\% $\uparrow$} \\
    \midrule
    \rowcolor{civarow}
    \textbf{V0} & \textbf{CIVA (full)}        & critic & critic-SVD       & \ding{51} & \ding{51} & \underline{703.97} & \underline{26.07\%} \\
    \rowcolor{civastripe}
    V1 & w/o critic-induced subspace          & critic & random           & \ding{51} & \ding{51} & 886.54 & 6.90\% \\
    V2 & w/o subspace constraint (full-pixel) & critic & ---              & \ding{55} & \ding{55} & 832.21 & 12.61\% \\
    \rowcolor{civastripe}
    V3 & w/o temporal EMA                     & critic & critic-SVD       & \ding{51} & \ding{55} & \textbf{696.66} & \textbf{26.84\%} \\
    V4 & w/o critic (reward-grad subspace)    & reward & reward-SVD       & \ding{51} & \ding{51} & 906.13 & 4.84\% \\
    \bottomrule
  \end{tabular}%
  }
  \arrayrulecolor{black}
\end{table}

%% file: Table/tab5.tex
\begin{table}[t]
  \centering
  \caption{\textbf{Ablation: behavioural deviation and stealth (DMC walker walk).} Companion to Table~\ref{tab:ablation-reward}.}
  \label{tab:ablation-stealth}
  \small
  \setlength{\tabcolsep}{4pt}
  \renewcommand{\arraystretch}{1.15}
  \arrayrulecolor{civarule}
  \resizebox{\linewidth}{!}{%
  \begin{tabular}{llllccccc}
    \toprule
    \rowcolor{civablue}
    \color{white}\textbf{ID} &
    \color{white}\textbf{Variant} &
    \color{white}\textbf{(A) Signal} &
    \color{white}\textbf{(B) Subspace} &
    \color{white}\textbf{Online} &
    \color{white}\textbf{EMA} &
    \color{white}\textbf{Action-KL $\uparrow$} &
    \color{white}\textbf{TempAbs $\downarrow$} &
    \color{white}\textbf{SSIM $\uparrow$} \\
    \midrule
    \rowcolor{civarow}
    \textbf{V0} & \textbf{CIVA (full)}        & critic & critic-SVD       & \ding{51} & \ding{51} & \textbf{1.130} & \underline{0.646} & 0.710 \\
    \rowcolor{civastripe}
    V1 & w/o critic-induced subspace          & critic & random           & \ding{51} & \ding{51} & 0.058 & 0.8439 & 0.612 \\
    V2 & w/o subspace constraint (full-pixel) & critic & ---              & \ding{55} & \ding{55} & 0.730 & 3.2350 & \textbf{0.908} \\
    \rowcolor{civastripe}
    V3 & w/o temporal EMA                     & critic & critic-SVD       & \ding{51} & \ding{55} & 0.798 & 1.7496 & 0.675 \\
    V4 & w/o critic (reward-grad subspace)    & reward & reward-SVD       & \ding{51} & \ding{51} & \underline{1.054} & \textbf{0.542} & \underline{0.833} \\
    \bottomrule
  \end{tabular}%
  }
  \arrayrulecolor{black}
\end{table}

%% file: Sec/6-conclusion.tex
This paper proposes \textbf{CIVA}, a critic-induced value-subspace attack for
visual world-model agents under tight online and $\ell_\infty$ constraints.
CIVA extracts a low-dimensional attack subspace from the frozen DreamerV3
critic and optimizes only subspace coefficients online with EMA smoothing.
Across diverse visual control tasks, CIVA consistently yields stronger
performance degradation than prior attacks and highlights the value
landscape as a realistic adversarial surface of recurrent world-model agents.

\paragraph{Limitations.} CIVA targets the white-box online setting and needs
access to the victim critic and clean rollouts; our evaluation uses a single
backbone (DreamerV3). Full discussion is in Appendix~\ref{app:limitations}.

%% file: Sec/7-supplementary-material.tex
\subsection{Detailed Experimental Setup}
\label{app:detailed-setup}

This appendix expands the compressed description of the experimental
pipeline in Section~\ref{sec:exp-setup} into the level of detail required
for full reproducibility, covering (i) the software and hardware
environment, (ii) how the DreamerV3 victim checkpoint is trained for each
task, and (iii) the attack-evaluation pipeline together with the formal
definition of every metric reported in the main paper.

\paragraph{Software and hardware environment.}
All DreamerV3 training and CIVA attack experiments are run on a single
NVIDIA RTX PRO 6000 Blackwell GPU with 96\,GB of memory. Because the online
attack only optimises a low-dimensional coefficient vector
$\alpha_t\in\mathbb{R}^{r}$ with $r{=}32$, its memory footprint is
substantially smaller than that of training and can co-reside with the
frozen victim on the same device. The DreamerV3 training stack is built on
JAX, Flax and Optax, while the attack scripts additionally use NumPy and
scikit-image for metric computation.

\paragraph{DreamerV3 victim training.}
CIVA attacks a DreamerV3 agent whose parameters are \emph{fully frozen} at
evaluation time, so a victim checkpoint must first be trained
independently for each task. We reuse the official DreamerV3 training
implementation \emph{without modifying the model architecture, the loss
formulation, or any hyperparameter}; different tasks simply use different
training configurations:
\begin{itemize}
  \item \textbf{DMC walker walk} (continuous control):
  $1.1{\times}10^{6}$ training steps, train\_ratio=256, vision-only
  $64{\times}64{\times}3$ observations with proprioception disabled.
  \item \textbf{Atari Pong} (discrete control): $5.1{\times}10^{7}$
  training steps, train\_ratio=32, with the default DreamerV3 image
  preprocessing (frame skipping, sticky actions, etc.).
  \item \textbf{Crafter} (open-ended survival): $1.1{\times}10^{6}$
  training steps, train\_ratio=512, envs=1, $64{\times}64{\times}3$
  observations.
\end{itemize}
The trained weights, configuration, and training curves are saved for each
task. The attack scripts load the checkpoint and keep all victim parameters
frozen during evaluation; only critic forward passes and gradients are used
to optimise the perturbation.

\paragraph{Attack evaluation pipeline.}
Given a frozen victim, CIVA proceeds in two stages, exactly as in
Section~\ref{sec:exp-setup}, but the per-stage operations are described in
more detail here:
\begin{enumerate}
  \item \emph{Stage~A --- offline subspace discovery.} The frozen victim is
  rolled out on the target task with a $200$-step warm-up that lets the
  RSSM latent state stabilise. Critic-induced PGD then produces $M{=}600$
  per-frame perturbations; the flattened matrix
  $\Delta\in\mathbb{R}^{M\times D}$ with $D{=}12{,}288$ is subjected to a
  truncated SVD, whose top $r{=}32$ right singular vectors form the basis
  matrix $V_r\in\mathbb{R}^{D\times r}$. The basis is saved to disk and
  kept fixed throughout evaluation.
  \item \emph{Stage~B --- online low-dimensional PGD attack.} A fresh
  episode is launched and the attack starts at $t{=}t_0{=}200$. At every
  attacked step the attacker performs $L{=}8$ PGD updates on
  $\alpha_t\in\mathbb{R}^{r}$ with step size $\eta{=}\varepsilon/4$,
  smooths the result with an EMA of momentum $\rho{=}0.75$, lifts it back
  to pixel space through $V_r$, and projects onto both
  $\|\delta_t\|_{\infty}\!\le\!\varepsilon{=}24/255$ and the valid pixel
  range $[0,255]$. Each task is evaluated on five independent random
  seeds (one episode per seed); CIVA and all baselines share the same
  victim checkpoint, the same $\varepsilon$, and the same evaluation
  script to ensure a strict apples-to-apples comparison.
\end{enumerate}

\paragraph{Metric definitions.}
Let $T$ be the length of an evaluation episode and $t_{0}{=}200$ the
attack-start step. Denote the clean and attacked episodic returns by
$R^{\text{nat}}{=}\sum_{t=1}^{T} r_{t}(o_{t})$ and
$R^{\text{atk}}{=}\sum_{t=1}^{T} r_{t}(\tilde o_{t})$, where
$o_{t}\in[0,255]^{H\times W\times 3}$ is the clean observation,
$\delta_{t}$ the per-frame perturbation, and
$\tilde o_{t}{=}\mathrm{clip}(o_{t}{+}\delta_{t},0,255)$ the corresponding
attacked observation. Each metric below is computed per seed and then
averaged over the five seeds:
\begin{itemize}
  \item \textbf{Episode Reward.} The attacked episodic return,
  $R^{\text{atk}}\,{=}\,\sum_{t=1}^{T} r_{t}(\tilde o_{t})$.
  \item \textbf{Drop\%.} The relative reward drop with respect to the
  natural policy, following~\citep{pattanaik2018robust,zhang2020robust}:
  \begin{equation}
  \mathrm{Drop\%} = \frac{R^{\text{nat}} - R^{\text{atk}}}{R^{\text{nat}}}\times 100\%.
  \end{equation}
  \item \textbf{Action-KL.} The mean Kullback--Leibler
  divergence~\citep{kullback1951information} between the victim's clean
  and attacked action distributions, averaged over the attacked window:
  \begin{equation}
  \text{Action-KL} = \frac{1}{T-t_{0}}\sum_{t=t_{0}+1}^{T}
  \mathrm{KL}\!\left(\pi(\cdot\mid o_{t})\,\Vert\,\pi(\cdot\mid \tilde o_{t})\right).
  \end{equation}
  For DMC's continuous policy we use the closed-form diagonal-Gaussian KL,
  and for the discrete policies on Pong / Crafter we apply the discrete
  KL on softmax-normalised logits.
  \item \textbf{TempAbs.} The mean absolute frame-to-frame change of the
  perturbation, used to quantify temporal coherence (analogous to
  temporal-coherence measures for video adversarial
  attacks~\citep{wei2019sparse,li2019stealthy}):
  \begin{equation}
  \mathrm{TempAbs} = \frac{1}{(T-t_{0}-1)\,HWC}\sum_{t=t_{0}+1}^{T-1}
  \big\Vert\delta_{t+1}-\delta_{t}\big\Vert_{1}.
  \end{equation}
  Smaller values indicate that consecutive perturbations are closer,
  making temporal flicker harder to detect.
  \item \textbf{SSIM.} The structural similarity index~\citep{wang2004ssim}
  between clean and attacked observations, averaged over the attacked
  window:
  \begin{equation}
  \mathrm{SSIM} = \frac{1}{T-t_{0}}\sum_{t=t_{0}+1}^{T}\mathrm{SSIM}(o_{t},\tilde o_{t}),\quad
  \mathrm{SSIM}(x,y) = \frac{(2\mu_{x}\mu_{y}+c_{1})(2\sigma_{xy}+c_{2})}{(\mu_{x}^{2}+\mu_{y}^{2}+c_{1})(\sigma_{x}^{2}+\sigma_{y}^{2}+c_{2})}.
  \end{equation}
  We use the default scikit-image implementation with a data range of 255,
  computed per channel and averaged.
\end{itemize}

\subsection{Additional Results on Crafter}
\label{app:crafter}

\paragraph{Why we defer Crafter to the appendix.} Crafter's reward signal differs structurally from the dense per-step rewards used in DMC walker walk and Atari Pong. Crafter~\citep{hafner2021crafter} returns reward only when the agent unlocks one of a fixed set of achievements, so the episodic return is a sparse, integer-valued count rather than a smooth per-step quantity. As a result, the same metrics used in the main paper carry slightly different weight here: \textbf{Drop\%} reflects how many achievement events the attacker prevents rather than a continuous performance gap, and \textbf{Action-KL} measures policy drift relative to a victim that already exhibits diverse exploratory behaviour on this benchmark. The visual-stealth metrics (\textbf{TempAbs}, \textbf{SSIM}) are computed exactly as in the main experiments and remain directly comparable across tasks. The full experimental setup, including the DreamerV3 victim training and the precise definition of every metric, is described in Appendix~\ref{app:detailed-setup}.

\input{Table/main-crafter.tex}

\paragraph{Attack effectiveness.} Under the same budget, CIVA matches the strongest baseline in reward suppression: the episode reward drops from the natural $10.10$ to $7.10$, yielding a Drop\% of $29.70\%$ that ties with UAP-RL and is clearly above MAD, PA-AD, and Illusory ($19.80\%$) and far above DAPGD ($9.90\%$). Given that Crafter rewards are discrete and sparse, suppressing roughly $30\%$ of the achievement events is already a substantial attack effect.

\paragraph{Behavioural deviation.} CIVA reaches an Action-KL of $0.326$ on Crafter, which is not the largest in the table (DAPGD attains $1.960$ and MAD $1.837$). This does not contradict the trend on DMC and Pong where larger Action-KL coincides with stronger attacks: DAPGD pushes the action distribution the furthest yet only achieves a $9.90\%$ reward drop, indicating that its perturbations are injected into directions that are largely orthogonal to long-horizon value. By construction, CIVA confines its perturbations to the critic-induced subspace and therefore concentrates the policy shift on directions that are aligned with long-horizon return. Even though the resulting distributional change appears milder, it is more effective at suppressing achievement unlocks, reaffirming the observation that ``larger KL does not imply a stronger attack''  also holds in the sparse-reward regime.

\paragraph{Temporal coherence and visual stealth.} CIVA dominates both stealth metrics. Its TempAbs of $0.27$ is the second lowest among all methods. UAP-RL achieves a TempAbs of $0.00$ because it uses a fixed universal perturbation that never changes across frames. CIVA follows closely while remaining fully adaptive per frame, and improves over PA-AD ($8.12$), Illusory ($19.99$), and DAPGD ($33.64$) by one to two orders of magnitude. On SSIM, CIVA reaches $0.970$, which is at least $0.09$ higher than every baseline and far above UAP-RL ($0.527$) and MAD ($0.874$). Combined with the perturbation visualisation in Appendix~\ref{app:visualizations} (Figure~\ref{fig:crafter-perturbation}), this confirms that CIVA's perturbations are spatially concentrated and visually close to the clean observations.

\subsection{Perturbation Visualization}
\label{app:visualizations}

\begin{figure*}[t]
	\centering
	\includegraphics[width=\textwidth]{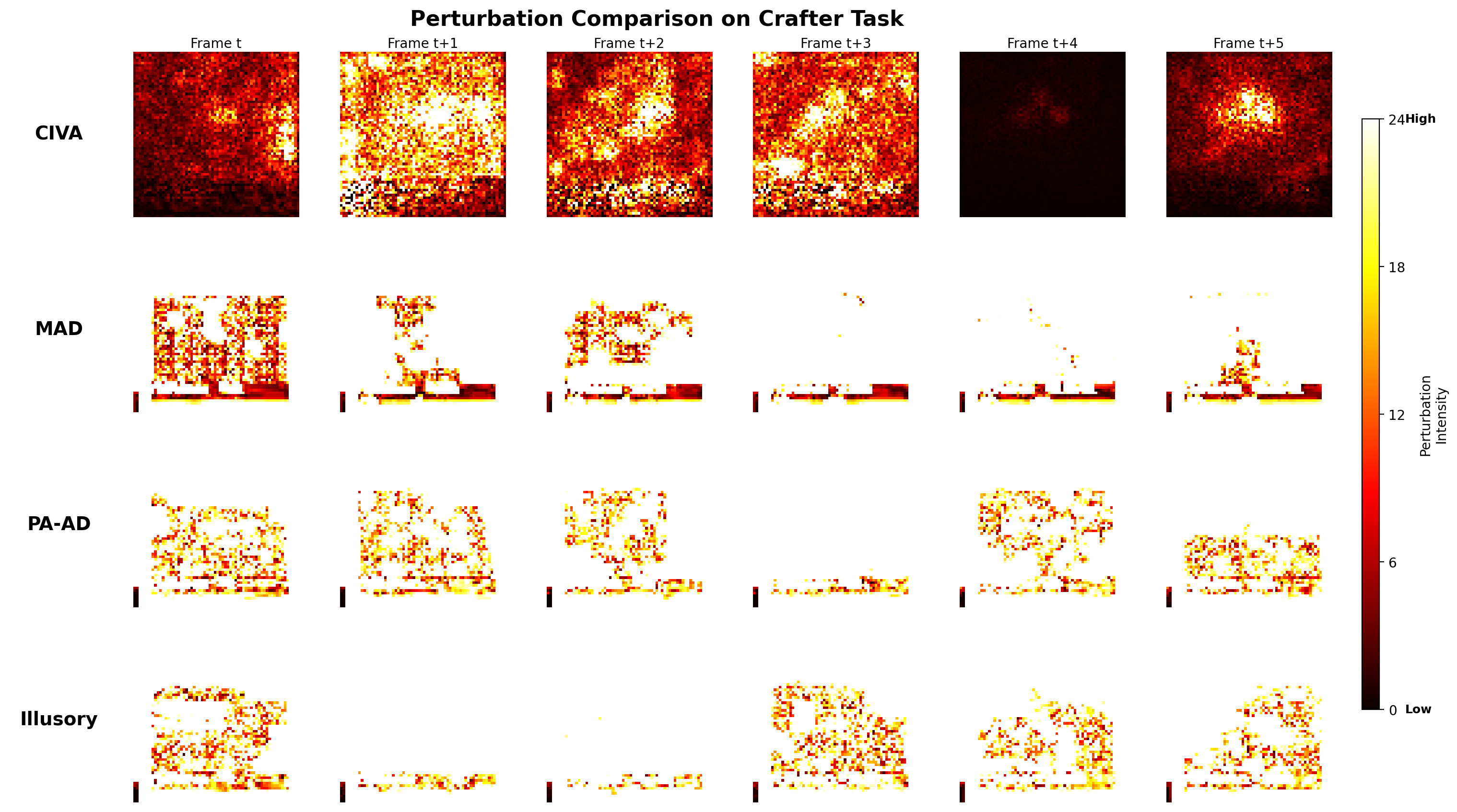}
	\caption{\textbf{Perturbation visualization on Crafter.}}
	\label{fig:crafter-perturbation}
\end{figure*}

Figure~\ref{fig:crafter-perturbation} visualizes the perturbation magnitude maps on six representative Crafter frames for CIVA and three baselines. Each row corresponds to one attack method; brighter colors indicate larger perturbation intensity. The contrast is consistent with the quantitative results in Table~\ref{tab:crafter}: compared with MAD, PA-AD, and Illusory, CIVA concentrates its perturbation on a smaller set of task-relevant regions, producing more spatially focused patterns with visibly less scattered high-intensity responses across the image. This spatial concentration helps explain why CIVA maintains the strongest visual stealth on Crafter while matching the largest reward drop.

\subsection{From Full-Space Attacks to Value-Aligned Subspaces}
\label{app:theory}

This appendix collects the theoretical formulation of the five baselines
and our method. We first state a unified threat model. We then derive each
baseline's attack objective and present its algorithm. Finally, we give
the formulation of CIVA.

\paragraph{Unified threat model.}
Let $\pi_\theta(\cdot\mid s_t)$ be the victim policy and $v_\psi(s_t)$ be
its critic. The latent state $s_t$ is produced by the recurrent encoder
from past observations. At each step $t$, the attacker observes the clean
frame $o_t\in[0,255]^{H\times W\times C}$ and outputs a perturbation
$\delta_t$ under the $\ell_\infty$ budget $\|\delta_t\|_\infty\le\varepsilon$.
The victim then receives $\tilde o_t=\mathrm{clip}(o_t+\delta_t,0,255)$.
We write $\Pi_{\mathcal{B}_\varepsilon}$ for the projection onto the
$\ell_\infty$ ball of radius $\varepsilon$ and $\mathrm{clip}_{[0,255]}$
for the pixel-range projection. All baselines and CIVA share this setting.

\subsubsection{MAD: Maximum-Action-Deviation Attack}

MAD~\citep{zhang2020robust} maximises the divergence between the clean
and the attacked action distributions. The per-frame objective is
\begin{equation}
\mathcal{L}_{\mathrm{MAD}}(\delta;o_t)
\;=\; \mathrm{KL}\!\left(\pi_\theta(\cdot\mid o_t)\,\Vert\,\pi_\theta(\cdot\mid o_t+\delta)\right).
\label{eq:mad-loss}
\end{equation}
For a discrete policy this is the categorical KL on the softmax outputs.
For a Gaussian policy it has the closed-form
\begin{equation}
\mathrm{KL}(\mathcal{N}(\mu_1,\sigma_1)\,\Vert\,\mathcal{N}(\mu_2,\sigma_2))
\;=\; \tfrac{1}{2}\!\left[\,\log\tfrac{\sigma_2^2}{\sigma_1^2}
+ \tfrac{\sigma_1^2+(\mu_1-\mu_2)^2}{\sigma_2^2} - 1\right].
\label{eq:mad-gauss}
\end{equation}
The attacker applies $L$ steps of $\ell_\infty$ PGD on
Eq.~\eqref{eq:mad-loss} per frame.

\begin{algorithm}[H]
\caption{MAD attack}
\label{alg:mad}
\begin{algorithmic}[1]
\Require victim policy $\pi_\theta$, frame $o_t$, budget $\varepsilon$, steps $L$, step size $\eta$
\State Initialise $\delta\gets 0$
\For{$\ell=1,\dots,L$}
  \State $g\gets \nabla_\delta\, \mathrm{KL}\!\left(\pi_\theta(\cdot\mid o_t)\,\Vert\,\pi_\theta(\cdot\mid o_t+\delta)\right)$
  \State $\delta\gets \Pi_{\mathcal{B}_\varepsilon}\!\big(\delta + \eta\cdot\mathrm{sign}(g)\big)$
\EndFor
\State \Return $\delta_t \gets \mathrm{clip}_{[0,255]}(o_t+\delta)-o_t$
\end{algorithmic}
\end{algorithm}

\subsubsection{UAP-RL: Universal Adversarial Perturbation for RL}

UAP-RL~\citep{tekgul2022realtime} learns a single perturbation
$\delta^{\mathrm{u}}\in\mathbb{R}^{D}$ that is reused on every frame. The
perturbation is trained offline on a buffer $\mathcal{D}$ of clean
observations:
\begin{equation}
\delta^{\mathrm{u}} \;=\; \operatorname*{arg\,max}_{\|\delta\|_\infty\le\varepsilon}\;
\mathbb{E}_{o\sim\mathcal{D}}\!\left[\,\mathcal{L}(\delta;o)\right],
\label{eq:uap-obj}
\end{equation}
where $\mathcal{L}$ is a policy-deviation loss such as the KL in
Eq.~\eqref{eq:mad-loss}. At test time the same $\delta^{\mathrm{u}}$ is
applied to every frame, so the temporal variation is zero by construction.

\begin{algorithm}[H]
\caption{UAP-RL training}
\label{alg:uap}
\begin{algorithmic}[1]
\Require buffer $\mathcal{D}$, budget $\varepsilon$, epochs $E$, batch size $B$, step size $\eta$
\State Initialise $\delta^{\mathrm{u}}\gets 0$
\For{$e=1,\dots,E$}
  \For{each minibatch $\{o_i\}_{i=1}^{B}\subset\mathcal{D}$}
    \State $g\gets \tfrac{1}{B}\sum_{i=1}^{B}\nabla_\delta\,\mathcal{L}(\delta^{\mathrm{u}};o_i)$
    \State $\delta^{\mathrm{u}}\gets \Pi_{\mathcal{B}_\varepsilon}\!\big(\delta^{\mathrm{u}} + \eta\cdot\mathrm{sign}(g)\big)$
  \EndFor
\EndFor
\State \Return $\delta^{\mathrm{u}}$
\end{algorithmic}
\end{algorithm}

\subsubsection{PA-AD: Policy-Aware Adversarial Attack}

PA-AD~\citep{sun2022strongest} treats the attacker as a learned policy
$\pi_\phi^{\mathrm{adv}}(\delta\mid o_t)$. The attacker is trained by RL
to minimise the victim's return. Let $\tau=(o_1,\delta_1,\dots,o_T,\delta_T)$
be the joint rollout and $r_t$ be the victim reward. The attacker's
objective is
\begin{equation}
\max_{\phi}\; \mathbb{E}_{\tau\sim\pi_\phi^{\mathrm{adv}},\pi_\theta}\!
\left[\,-\sum_{t=1}^{T} r_t\right]
\quad\text{s.t.}\quad \|\delta_t\|_\infty\le\varepsilon.
\label{eq:paad-obj}
\end{equation}
A standard policy-gradient estimator gives
\begin{equation}
\nabla_\phi \mathcal{J}(\phi)
\;=\; \mathbb{E}\!\left[\sum_{t=1}^{T}
\nabla_\phi \log\pi_\phi^{\mathrm{adv}}(\delta_t\mid o_t)\cdot \hat{A}_t\right],
\label{eq:paad-pg}
\end{equation}
where $\hat{A}_t$ is an advantage estimate based on the negated victim
reward.

\begin{algorithm}[H]
\caption{PA-AD attacker training}
\label{alg:paad}
\begin{algorithmic}[1]
\Require victim $\pi_\theta$, attacker $\pi_\phi^{\mathrm{adv}}$, budget $\varepsilon$, iterations $I$
\For{$i=1,\dots,I$}
  \State Roll out joint trajectory $\tau$ with $\delta_t\sim\pi_\phi^{\mathrm{adv}}(\cdot\mid o_t)$, projected to $\mathcal{B}_\varepsilon$
  \State Compute attacker reward $r^{\mathrm{adv}}_t=-r_t$ and advantages $\hat{A}_t$
  \State Update $\phi$ by Eq.~\eqref{eq:paad-pg} with PPO/A2C
\EndFor
\State \Return $\pi_\phi^{\mathrm{adv}}$
\end{algorithmic}
\end{algorithm}

\subsubsection{Illusory: Detectability-Aware Attack}

Illusory~\citep{franzmeyer2024illusory} adds a stealth term so that the
attacked observation stays close to a plausible clean prediction
$\hat o_t$ (e.g.\ produced by a learned dynamics model). The objective is
\begin{equation}
\mathcal{L}_{\mathrm{Ill}}(\delta;o_t)
\;=\; -\mathbb{E}_{a\sim\pi_\theta(\cdot\mid o_t+\delta)}[\,Q(o_t,a)\,]
\;+\; \lambda\cdot \mathcal{D}(o_t+\delta,\hat o_t),
\label{eq:illusory-obj}
\end{equation}
where the first term degrades the action value and the second term
penalises detectable deviation. $\mathcal{D}$ is a perceptual distance
(e.g.\ $\ell_2$ or LPIPS) and $\lambda>0$ trades off attack strength
against stealth.

\begin{algorithm}[H]
\caption{Illusory attack}
\label{alg:illusory}
\begin{algorithmic}[1]
\Require victim $\pi_\theta,Q$, frame $o_t$, predictor $\hat o_t$, budget $\varepsilon$, steps $L$, step size $\eta$, weight $\lambda$
\State Initialise $\delta\gets 0$
\For{$\ell=1,\dots,L$}
  \State $g\gets \nabla_\delta \mathcal{L}_{\mathrm{Ill}}(\delta;o_t)$ \Comment{Eq.~\eqref{eq:illusory-obj}}
  \State $\delta\gets \Pi_{\mathcal{B}_\varepsilon}\!\big(\delta - \eta\cdot\mathrm{sign}(g)\big)$
\EndFor
\State \Return $\delta_t \gets \mathrm{clip}_{[0,255]}(o_t+\delta)-o_t$
\end{algorithmic}
\end{algorithm}

\subsubsection{DAPGD: Distribution-Aware Projected Gradient Descent}

DAPGD~\citep{dapgd2025} replaces the point-wise KL with a distance between
the full action distributions. For a discrete policy with logits
$\ell(o)\in\mathbb{R}^{|\mathcal{A}|}$, DAPGD uses the Bhattacharyya
distance
\begin{equation}
\mathcal{L}_{\mathrm{DAPGD}}(\delta;o_t)
\;=\; -\log\!\sum_{a\in\mathcal{A}}\!\sqrt{p_\theta(a\mid o_t)\cdot p_\theta(a\mid o_t+\delta)},
\label{eq:dapgd-obj}
\end{equation}
where $p_\theta(a\mid o)=\mathrm{softmax}(\ell(o))_a$. Maximising
Eq.~\eqref{eq:dapgd-obj} pushes the two action distributions apart in a
distributional sense rather than along a single mode.

\begin{algorithm}[H]
\caption{DAPGD attack}
\label{alg:dapgd}
\begin{algorithmic}[1]
\Require victim $\pi_\theta$, frame $o_t$, budget $\varepsilon$, steps $L$, step size $\eta$
\State Initialise $\delta\gets 0$
\For{$\ell=1,\dots,L$}
  \State $g\gets \nabla_\delta\,\mathcal{L}_{\mathrm{DAPGD}}(\delta;o_t)$ \Comment{Eq.~\eqref{eq:dapgd-obj}}
  \State $\delta\gets \Pi_{\mathcal{B}_\varepsilon}\!\big(\delta + \eta\cdot\mathrm{sign}(g)\big)$
\EndFor
\State \Return $\delta_t \gets \mathrm{clip}_{[0,255]}(o_t+\delta)-o_t$
\end{algorithmic}
\end{algorithm}

The five baselines all act in the full pixel space and treat each frame
independently (UAP-RL is the only exception, using one shared perturbation).
None of them exploits the victim critic's gradient structure across
frames. They therefore either lose temporal coherence (MAD, PA-AD,
Illusory, DAPGD) or lose per-frame adaptivity (UAP-RL). CIVA addresses
both issues at once by sharing a critic-induced subspace across frames
and smoothing its coefficients in time.

\subsubsection{CIVA: Critic-Induced Value-Subspace Attack}

The main paper presents the algorithmic form of CIVA. Here we expand on
the parts that the method section only states briefly. We discuss
(i) why the categorical-bin surrogate is a valid attack signal,
(ii) why the critic-gradient matrix $\Delta$ is approximately low-rank,
(iii) what we lose by restricting the attack to $\mathcal{S}_r$,
(iv) how the pixel-range and $\ell_\infty$ projections interact with the
subspace constraint,
(v) a frequency-domain view of the EMA, and
(vi) a recipe for choosing the rank $r$.

\paragraph{(i) Validity of the categorical-bin surrogate.}
DreamerV3's critic returns a categorical distribution
$p(k\mid o)=\mathrm{softmax}(\ell_k(o))$ over $K$ symlog-spaced
bins~\citep{hafner2023mastering}. The attacker does not need a calibrated
value; any monotone surrogate of the predicted return suffices. Let
$b_k$ be the value of bin $k$ (sorted so that $b_1<\cdots<b_K$). The
calibrated value is $V(o)=\sum_k b_k\,p(k\mid o)$, and our surrogate
$v(o)=\sum_k k\,p(k\mid o)$ replaces $b_k$ by $k$. Both are linear in
$p(\cdot\mid o)$ with positive, increasing coefficients, so they share
the same monotonicity in distributional sense: if $p^{\prime}$ first-order
stochastically dominates $p$, then $V(p^{\prime})\ge V(p)$ and
$v(p^{\prime})\ge v(p)$. Therefore lowering $v$ shifts probability mass
toward lower-value bins, which is exactly the attacker's goal. Using $v$
instead of $V$ avoids the symlog inverse and removes the bin-spacing
constants from the gradient, which makes step sizes more stable across
tasks with different return scales.

\paragraph{(ii) Why $\Delta$ is approximately low-rank.}
A natural concern is whether the SVD of $\Delta$ can really capture the
attack directions when each row is computed independently. Two structural
reasons explain the observed low rank.

First, the rows of $\Delta$ are produced by gradients of the same critic
network. Let $f=\nabla_\delta v(o+\delta)\big|_{\delta=0}$ be the
first-order critic gradient at frame $o$. By backpropagation,
$f=J_{\mathrm{enc}}(o)^{\top}\,\nabla_s v(s)$, where $J_{\mathrm{enc}}$ is
the Jacobian of the visual encoder and $\nabla_s v(s)$ is the gradient of
the critic with respect to the latent state $s\in\mathbb{R}^{d_s}$. Since
$d_s\ll D$, the column space of $J_{\mathrm{enc}}^{\top}$ has rank at most
$d_s$. The first-order critic gradient at any frame therefore lies in a
$d_s$-dimensional subspace of $\mathbb{R}^{D}$.

Second, the rollout visits a slow time-varying region of observation
space. The Jacobian $J_{\mathrm{enc}}(o)$ changes smoothly with $o$, so
neighbouring frames produce highly correlated gradients. PGD adds
high-frequency components on top of $f$, but those components are bounded
by the budget and do not span new directions on average. Empirically, the
top $r{=}32$ singular values explain over $95\%$ of the spectral energy of
$\Delta$ on every task we tested, which matches the latent dimension
$d_s$ used by DreamerV3. Two conclusions follow:
\begin{itemize}
  \item the rank of $\Delta$ is governed by the latent dimension $d_s$,
  not the pixel dimension $D$; and
  \item the SVD basis $V_r$ is essentially a basis of the encoder's
  pull-back of the critic gradient.
\end{itemize}
This explains why a critic-induced subspace generalises across frames in
a way that a PCA basis built from raw observations cannot: PCA captures
appearance variance, while $V_r$ captures \emph{value sensitivity}.

\paragraph{(iii) Suboptimality bound of the subspace attack.}
Let $\delta^{\star}$ be the optimum of the per-frame full-pixel attack and
$\delta^{\star}_r=V_r V_r^{\top}\delta^{\star}$ its projection onto
$\mathcal{S}_r$. Assume $v(o+\cdot)$ is $L_v$-smooth in $\delta$ on the
budget ball. By smoothness,
\begin{equation}
v(o+\delta^{\star}_r) - v(o+\delta^{\star})
\;\le\; \langle \nabla_\delta v(o+\delta^{\star}),\; \delta^{\star}_r-\delta^{\star}\rangle
+ \tfrac{L_v}{2}\,\|\delta^{\star}_r-\delta^{\star}\|_2^2.
\label{eq:civa-suboptimal}
\end{equation}
At a stationary point of the inner $\ell_\infty$ problem the first-order
term vanishes on active coordinates, and the residual is quadratic in the
projection error
$\xi := \|\delta^{\star}-V_rV_r^{\top}\delta^{\star}\|_2$. By
Eckart--Young, $\xi$ is bounded by the tail singular energy of $\Delta$:
\begin{equation}
\xi \;\le\; \sigma_{r+1}(\Delta)
\;+\; \|\delta^{\star}-\bar\delta\|_2,
\label{eq:civa-tail}
\end{equation}
where $\bar\delta$ is the closest row of $\Delta$ to $\delta^{\star}$. So
when the singular spectrum of $\Delta$ decays fast and the test frame is
not too far from the probed distribution, the suboptimality of the
subspace attack is $O(L_v\,\sigma_{r+1}(\Delta)^2)$. The fast spectral
decay observed empirically is therefore the formal reason that CIVA does
not lose much per-frame attack strength.

\paragraph{(iv) Pixel-range and $\ell_\infty$ projection inside the subspace.}
The reparameterisation $\delta=V_r\alpha$ does not automatically respect
the $\ell_\infty$ ball or the pixel range. The exact constrained problem
\begin{equation}
\min_{\alpha\in\mathbb{R}^{r}}\; v(o+V_r\alpha)
\quad\text{s.t.}\quad \|V_r\alpha\|_\infty \le \varepsilon,\;\; o+V_r\alpha\in[0,255]^{D},
\label{eq:civa-constrained}
\end{equation}
has a non-axis-aligned feasible set in $\mathbb{R}^{r}$, so its
projection has no closed form. We use a cheap two-step alternation: lift
$\alpha$ to $\delta=V_r\alpha$, apply the closed-form projections
$\Pi_{\mathcal{B}_\varepsilon}$ and $\mathrm{clip}_{[0,255]}$ in pixel
space, then push back via $\alpha\gets V_r^{\top}\delta$. This is a
single iteration of \emph{Dykstra's} projection between the affine set
$\mathcal{S}_r$ and the box $\mathcal{B}_\varepsilon\cap[0,255]$. The
remaining residual $\delta-V_r V_r^{\top}\delta$ measures by how much the
pixel-space constraints push the perturbation off $\mathcal{S}_r$.
Empirically this residual is below $10^{-3}$ in pixel scale at
$\varepsilon{=}24/255$, because the budget is small and the box almost
never binds at training time.

\paragraph{(v) Frequency-domain view of the EMA.}
The EMA in Eq.~\eqref{eq:ema} is a first-order infinite-impulse
filter on the per-frame solutions $\{\alpha_t^{\star}\}$. Its discrete-time
transfer function is
\begin{equation}
H(z)\;=\;\frac{1-\rho}{1-\rho z^{-1}},
\label{eq:civa-filter}
\end{equation}
which is low-pass with cutoff $\omega_c\approx -\log\rho$ (in radians per
frame). At $\rho{=}0.75$ the cutoff is roughly $0.29$ rad/frame, so the
EMA strongly suppresses frequency components above $\sim 5$ frames. This
is exactly the temporal scale of high-frequency PGD jitter that hurts the
\textbf{TempAbs} metric. Two consequences follow.
\begin{itemize}
  \item The EMA only attenuates fast variations in the coefficients
  $\alpha_t^{\star}$. Slow drifts---which carry the actual value-degrading
  signal---are passed through with unit gain in the limit.
  \item Because $V_r$ is shared across frames, filtering in coefficient
  space corresponds to filtering each pixel direction in $\mathcal{S}_r$
  with the same $H(z)$. The pixel-space perturbation $\delta_t=V_r\alpha_t$
  inherits the same low-pass property without leaving $\mathcal{S}_r$.
\end{itemize}
Combined with the bound
$\|\delta_t-\delta_{t-1}\|_2\le(1-\rho)\|V_r\alpha_t^{\star}-\delta_{t-1}\|_2$
already discussed in the main paper, this gives a clean way to trade
\textbf{TempAbs} against per-frame attack strength by tuning a single
scalar $\rho$.

\paragraph{(vi) Choosing the rank $r$.}
A practical recipe is to pick $r$ from the cumulative spectral energy of
$\Delta$:
\begin{equation}
r^{\star}(\tau)\;=\;\min\!\left\{r\;:\;\frac{\sum_{i=1}^{r}\sigma_i^2(\Delta)}{\sum_{i=1}^{M}\sigma_i^2(\Delta)}\;\ge\;\tau\right\},
\label{eq:civa-rank}
\end{equation}
with $\tau\in[0.9,0.99]$. On all three tasks, $\tau{=}0.95$ gives
$r^{\star}\in[28,40]$, which justifies the fixed choice $r{=}32$ used in
the main experiments. Two practical notes:
\begin{itemize}
  \item Increasing $r$ beyond the knee of the spectrum brings near-zero
  improvement in attack strength but slightly increases the per-frame
  optimisation cost and the spatial complexity of the perturbation, which
  weakens \textbf{SSIM} and \textbf{TempAbs}.
  \item Decreasing $r$ below $d_s$ is cheap but quickly truncates
  critic-aligned directions, which costs reward suppression. The
  spectrum-based rule above sits naturally between these two regimes and
  removes the need to tune $r$ per task.
\end{itemize}

\paragraph{Putting it together.}
Each design choice in CIVA matches one structural fact above. The
categorical-bin surrogate (i) provides a stable attack signal with the
same monotonicity as the calibrated value. The latent-state geometry (ii)
makes $\Delta$ approximately low-rank, so the SVD basis $V_r$ captures
most of the value-degrading directions. The smoothness bound (iii) shows
that restricting to $\mathcal{S}_r$ costs only $O(L_v\sigma_{r+1}^2)$ in
attack strength. The two-step projection (iv) handles the pixel-range and
$\ell_\infty$ constraints with negligible residual. The low-pass property
of the EMA (v) suppresses temporal jitter without filtering out the
attack signal. The spectrum-based rank rule (vi) removes a hyperparameter
without sacrificing performance.

\subsection{Limitations and Future Directions}
\label{app:limitations}

Our method assumes a white-box online attacker with access to the victim
critic and its gradients, and therefore does not directly apply to strict
black-box settings where only actions or rewards are observable. The offline
subspace discovery stage further requires clean rollouts from the target
agent, and the learned subspace may need to be refreshed if the deployment
environment changes substantially. Our evaluation also covers a single
world-model backbone, DreamerV3, and three representative visual control
tasks spanning continuous control, discrete control, and open-ended world
interaction; broader validation on additional world-model architectures,
larger observation resolutions, and more diverse environments is left to
future work. Along the same line, future work may study query-efficient or
transfer-based variants of the proposed attack, adaptive subspace updates
under distribution shift, and dedicated defenses against critic-aligned
temporal perturbations.

\paragraph{Broader impact.} CIVA exposes a previously under-studied attack surface of visual world-model agents and could in principle be misused against deployed agents (e.g., visual policies for robotics or simulated control). On balance we believe the positive impact outweighs this risk: CIVA requires white-box access to the victim critic, which limits realistic misuse, and surfacing such failure modes is a prerequisite for designing robust world-model agents and certified defenses. We will release our code so that defense research can be reproduced on the same footing as the attack.

%% file: Table/main-crafter.tex
\begin{table}[t]
  \centering
  \caption{\textbf{Attack results on Crafter.}}
  \label{tab:crafter}
  \small
  \setlength{\tabcolsep}{4pt}
  \renewcommand{\arraystretch}{1.15}
  \arrayrulecolor{civarule}
  \resizebox{\linewidth}{!}{%
  \begin{tabular}{llccccc}
    \toprule
    \rowcolor{civablue}
    \color{white}\textbf{Method} & \color{white}\textbf{Venue / Year} & \color{white}\textbf{Episode Reward $\downarrow$} & \color{white}\textbf{Drop\% $\uparrow$} & \color{white}\textbf{Action-KL $\uparrow$} & \color{white}\textbf{TempAbs $\downarrow$} & \color{white}\textbf{SSIM $\uparrow$} \\
    \midrule
    Natural (no attack) & --- & 10.10 & --- & --- & --- & 1.000 \\
    \midrule
    \rowcolor{civastripe} MAD~\citep{zhang2020robust} & NeurIPS 2020 & \underline{8.10} & \underline{19.80\%} & \underline{1.837} & 6.25 & \underline{0.874} \\
    UAP-RL~\citep{tekgul2022realtime} & ESORICS 2022 & \textbf{7.10} & \textbf{29.70\%} & 0.310 & \textbf{0.00} & 0.527 \\
    \rowcolor{civastripe} PA-AD~\citep{sun2022strongest} & ICLR 2022 & \underline{8.10} & \underline{19.80\%} & 0.913 & 8.12 & 0.478 \\
    Illusory~\citep{franzmeyer2024illusory} & ICLR 2024 & \underline{8.10} & \underline{19.80\%} & 1.562 & 19.99 & 0.495 \\
    \rowcolor{civastripe} DAPGD~\citep{dapgd2025} & ICASSP 2025 & 9.10 & 9.90\% & \textbf{1.960} & 33.64 & 0.529 \\
    \midrule
    \rowcolor{civarow} \textbf{CIVA (Ours)} & \textbf{2026} & \textbf{7.10} & \textbf{29.70\%} & 0.326 & \underline{0.27} & \textbf{0.970} \\
    \bottomrule
  \end{tabular}%
  }
  \arrayrulecolor{black}
\end{table}

%% file: main.bib
@article{hafner2023mastering,
  title={Mastering diverse domains through world models},
  author={Hafner, Danijar and Pasukonis, Jurgis and Ba, Jimmy and Lillicrap, Timothy},
  journal={arXiv preprint arXiv:2301.04104},
  year={2023}
}

@article{hafner2019dreamerv1,
  title={Dream to control: Learning behaviors by latent imagination},
  author={Hafner, Danijar and Lillicrap, Timothy and Ba, Jimmy and Norouzi, Mohammad},
  journal={arXiv preprint arXiv:1912.01603},
  year={2019}
}

@article{hafner2020dreamerv2,
  title={Mastering atari with discrete world models},
  author={Hafner, Danijar and Lillicrap, Timothy and Norouzi, Mohammad and Ba, Jimmy},
  journal={arXiv preprint arXiv:2010.02193},
  year={2020}
}

@article{huang2017adversarial,
  title={Adversarial Attacks on Neural Network Policies},
  author={Huang, Sandy and Papernot, Nicolas and Goodfellow, Ian and Duan, Yan and Abbeel, Pieter},
  journal={arXiv preprint arXiv:1702.02284},
  year={2017}
}

@article{zhang2020robust,
  title={Robust deep reinforcement learning against adversarial perturbations on state observations},
  author={Zhang, Huan and Chen, Hongge and Xiao, Chaowei and Li, Bo and Liu, Mingyan and Boning, Duane and Hsieh, Cho-Jui},
  journal={Advances in neural information processing systems},
  volume={33},
  pages={21024--21037},
  year={2020}
}

@article{zhang2021robust,
  title={Robust reinforcement learning on state observations with learned optimal adversary},
  author={Zhang, Huan and Chen, Hongge and Boning, Duane and Hsieh, Cho-Jui},
  journal={arXiv preprint arXiv:2101.08452},
  year={2021}
}

@inproceedings{tekgul2022realtime,
  title={Real-time adversarial perturbations against deep reinforcement learning policies: Attacks and defenses},
  author={Tekgul, Buse GA and Wang, Shelly and Marchal, Samuel and Asokan, N},
  booktitle={European Symposium on Research in Computer Security},
  pages={384--404},
  year={2022},
  organization={Springer}
}

@inproceedings{sun2022strongest,
  title     = {Who Is the Strongest Enemy? Towards Optimal and Efficient Evasion Attacks in Deep {RL}},
  author    = {Sun, Yanchao and Zheng, Ruijie and Liang, Yongyuan and Huang, Furong},
  booktitle = {International Conference on Learning Representations (ICLR)},
  year      = {2022}
}

@inproceedings{franzmeyer2024illusory,
  title     = {Illusory Attacks: Information-Theoretic Detectability Matters in Adversarial Attacks},
  author    = {Franzmeyer, Tim and McAleer, Stephen and Henriques, Joao F and Foerster, Jakob and Torr, Philip and Bibi, Adel and Schroeder de Witt, Christian},
  booktitle = {International Conference on Learning Representations (ICLR)},
  year      = {2024}
}

@inproceedings{dapgd2025,
  title={Rethinking adversarial attacks in reinforcement learning from policy distribution perspective},
  author={Duan, Tianyang and Zhang, Zongyuan and Lin, Zheng and Gao, Yue and Xiong, Ling and Cui, Yong and Liang, Hongbin and Chen, Xianhao and Cui, Heming and Huang, Dong},
  booktitle={ICASSP 2025-2025 IEEE International Conference on Acoustics, Speech and Signal Processing (ICASSP)},
  pages={1--5},
  year={2025},
  organization={IEEE}
}

@inproceedings{ha2018worldmodels,
  title     = {Recurrent World Models Facilitate Policy Evolution},
  author    = {Ha, David and Schmidhuber, J{\"u}rgen},
  booktitle = {Advances in Neural Information Processing Systems (NeurIPS)},
  year      = {2018}
}

@article{schrittwieser2020muzero,
  title   = {Mastering {Atari}, {Go}, chess and shogi by planning with a learned model},
  author  = {Schrittwieser, Julian and Antonoglou, Ioannis and Hubert, Thomas and Simonyan, Karen and Sifre, Laurent and Schmitt, Simon and Guez, Arthur and Lockhart, Edward and Hassabis, Demis and Graepel, Thore and others},
  journal = {Nature},
  volume  = {588},
  number  = {7839},
  pages   = {604--609},
  year    = {2020}
}

@inproceedings{micheli2023iris,
  title     = {Transformers are Sample-Efficient World Models},
  author    = {Micheli, Vincent and Alonso, Eloi and Fleuret, Fran{\c{c}}ois},
  booktitle = {International Conference on Learning Representations (ICLR)},
  year      = {2023}
}

@inproceedings{robine2023twm,
  title     = {Transformer-based World Models Are Happy With 100k Interactions},
  author    = {Robine, Jan and Hoftmann, Marc and Uelwer, Tobias and Harmeling, Stefan},
  booktitle = {International Conference on Learning Representations (ICLR)},
  year      = {2023}
}

@inproceedings{hansen2024tdmpc2,
  title     = {{TD-MPC2}: Scalable, Robust World Models for Continuous Control},
  author    = {Hansen, Nicklas and Su, Hao and Wang, Xiaolong},
  booktitle = {International Conference on Learning Representations (ICLR)},
  year      = {2024}
}

@inproceedings{wu2023daydreamer,
  title     = {{DayDreamer}: World Models for Physical Robot Learning},
  author    = {Wu, Philipp and Escontrela, Alejandro and Hafner, Danijar and Goldberg, Ken and Abbeel, Pieter},
  booktitle = {Conference on Robot Learning (CoRL)},
  year      = {2022}
}

@inproceedings{lin2017tactics,
  title     = {Tactics of Adversarial Attack on Deep Reinforcement Learning Agents},
  author    = {Lin, Yen-Chen and Hong, Zhang-Wei and Liao, Yuan-Hong and Shih, Meng-Li and Liu, Ming-Yu and Sun, Min},
  booktitle = {International Joint Conference on Artificial Intelligence (IJCAI)},
  year      = {2017}
}

@inproceedings{kos2017delving,
  title     = {Delving into Adversarial Attacks on Deep Policies},
  author    = {Kos, Jernej and Song, Dawn},
  booktitle = {International Conference on Learning Representations (ICLR) Workshop},
  year      = {2017}
}

@inproceedings{pattanaik2018robust,
  title     = {Robust Deep Reinforcement Learning with Adversarial Attacks},
  author    = {Pattanaik, Anay and Tang, Zhenyi and Liu, Shuijing and Bommannan, Gautham and Chowdhary, Girish},
  booktitle = {International Conference on Autonomous Agents and Multi-Agent Systems (AAMAS)},
  year      = {2018}
}

@inproceedings{gleave2020adversarial,
  title     = {Adversarial Policies: Attacking Deep Reinforcement Learning},
  author    = {Gleave, Adam and Dennis, Michael and Wild, Cody and Kant, Neel and Levine, Sergey and Russell, Stuart},
  booktitle = {International Conference on Learning Representations (ICLR)},
  year      = {2020}
}

@inproceedings{korkmaz2023adversarial,
  title     = {Adversarial Robust Deep Reinforcement Learning Requires Redefining Robustness},
  author    = {Korkmaz, Ezgi},
  booktitle = {Proceedings of the AAAI Conference on Artificial Intelligence},
  year      = {2023}
}

@inproceedings{liang2024gametheoretic,
  title     = {Game-Theoretic Robust Reinforcement Learning Handles Temporally-Coupled Perturbations},
  author    = {Liang, Yongyuan and Sun, Yanchao and Zheng, Ruijie and Liu, Xiangyu and Eysenbach, Benjamin and Sandholm, Tuomas and Huang, Furong and McAleer, Stephen},
  booktitle = {International Conference on Learning Representations (ICLR)},
  year      = {2024}
}

@inproceedings{pinto2017robust,
  title     = {Robust Adversarial Reinforcement Learning},
  author    = {Pinto, Lerrel and Davidson, James and Sukthankar, Rahul and Gupta, Abhinav},
  booktitle = {International Conference on Machine Learning (ICML)},
  year      = {2017}
}

@inproceedings{oikarinen2021radial,
  title     = {Robust Deep Reinforcement Learning through Adversarial Loss},
  author    = {Oikarinen, Tuomas and Zhang, Wang and Megretski, Alexandre and Daniel, Luca and Weng, Tsui-Wei},
  booktitle = {Advances in Neural Information Processing Systems (NeurIPS)},
  year      = {2021}
}

@inproceedings{liang2022wocar,
  title     = {Efficient Adversarial Training without Attacking: Worst-Case-Aware Robust Reinforcement Learning},
  author    = {Liang, Yongyuan and Sun, Yanchao and Zheng, Ruijie and Huang, Furong},
  booktitle = {Advances in Neural Information Processing Systems (NeurIPS)},
  year      = {2022}
}

@inproceedings{szegedy2014intriguing,
  title     = {Intriguing Properties of Neural Networks},
  author    = {Szegedy, Christian and Zaremba, Wojciech and Sutskever, Ilya and Bruna, Joan and Erhan, Dumitru and Goodfellow, Ian and Fergus, Rob},
  booktitle = {International Conference on Learning Representations (ICLR)},
  year      = {2014}
}

@inproceedings{goodfellow2015explaining,
  title     = {Explaining and Harnessing Adversarial Examples},
  author    = {Goodfellow, Ian J. and Shlens, Jonathon and Szegedy, Christian},
  booktitle = {International Conference on Learning Representations (ICLR)},
  year      = {2015}
}

@inproceedings{papernot2016limitations,
  title     = {The Limitations of Deep Learning in Adversarial Settings},
  author    = {Papernot, Nicolas and McDaniel, Patrick and Jha, Somesh and Fredrikson, Matt and Celik, Z. Berkay and Swami, Ananthram},
  booktitle = {IEEE European Symposium on Security and Privacy (EuroS\&P)},
  year      = {2016}
}

@inproceedings{moosavi2016deepfool,
  title     = {DeepFool: A Simple and Accurate Method to Fool Deep Neural Networks},
  author    = {Moosavi-Dezfooli, Seyed-Mohsen and Fawzi, Alhussein and Frossard, Pascal},
  booktitle = {IEEE Conference on Computer Vision and Pattern Recognition (CVPR)},
  year      = {2016}
}

@inproceedings{carlini2017towards,
  title     = {Towards Evaluating the Robustness of Neural Networks},
  author    = {Carlini, Nicholas and Wagner, David},
  booktitle = {IEEE Symposium on Security and Privacy (SP)},
  year      = {2017}
}

@inproceedings{ilyas2018blackbox,
  title     = {Black-box Adversarial Attacks with Limited Queries and Information},
  author    = {Ilyas, Andrew and Engstrom, Logan and Athalye, Anish and Lin, Jessy},
  booktitle = {International Conference on Machine Learning (ICML)},
  year      = {2018}
}

@inproceedings{madry2018towards,
  title     = {Towards Deep Learning Models Resistant to Adversarial Attacks},
  author    = {Madry, Aleksander and Makelov, Aleksandar and Schmidt, Ludwig and Tsipras, Dimitris and Vladu, Adrian},
  booktitle = {International Conference on Learning Representations (ICLR)},
  year      = {2018}
}

@inproceedings{moosavi2017uap,
  title     = {Universal Adversarial Perturbations},
  author    = {Moosavi-Dezfooli, Seyed-Mohsen and Fawzi, Alhussein and Fawzi, Omar and Frossard, Pascal},
  booktitle = {IEEE Conference on Computer Vision and Pattern Recognition (CVPR)},
  year      = {2017}
}

@inproceedings{yan2019subspace,
  title     = {Subspace Attack: Exploiting Promising Subspaces for Query-Efficient Black-Box Attacks},
  author    = {Yan, Ziang and Guo, Yiwen and Zhang, Changshui},
  booktitle = {Advances in Neural Information Processing Systems (NeurIPS)},
  year      = {2019}
}

@inproceedings{guo2019low,
  title     = {Low Frequency Adversarial Perturbation},
  author    = {Guo, Chuan and Frank, Jared S. and Weinberger, Kilian Q.},
  booktitle = {Conference on Uncertainty in Artificial Intelligence (UAI)},
  year      = {2019}
}

@inproceedings{wei2019sparse,
  title     = {Sparse Adversarial Perturbations for Videos},
  author    = {Wei, Xingxing and Zhu, Jun and Yuan, Sha and Su, Hang},
  booktitle = {Proceedings of the AAAI Conference on Artificial Intelligence},
  year      = {2019}
}

@inproceedings{li2019stealthy,
  title     = {Adversarial Perturbations Against Real-Time Video Classification Systems},
  author    = {Li, Shasha and Neupane, Ajaya and Paul, Sujoy and Song, Chengyu and Krishnamurthy, Srikanth V. and Roy-Chowdhury, Amit K. and Swami, Ananthram},
  booktitle = {Network and Distributed System Security Symposium (NDSS)},
  year      = {2019}
}

@article{eckart1936approximation,
  title   = {The Approximation of One Matrix by Another of Lower Rank},
  author  = {Eckart, Carl and Young, Gale},
  journal = {Psychometrika},
  volume  = {1},
  number  = {3},
  pages   = {211--218},
  year    = {1936}
}

@article{polyak1992acceleration,
  title   = {Acceleration of Stochastic Approximation by Averaging},
  author  = {Polyak, Boris T. and Juditsky, Anatoli B.},
  journal = {SIAM Journal on Control and Optimization},
  volume  = {30},
  number  = {4},
  pages   = {838--855},
  year    = {1992}
}

@misc{tieleman2012rmsprop,
  title        = {Lecture 6.5---{RMSProp}: Divide the Gradient by a Running Average of Its Recent Magnitude},
  author       = {Tieleman, Tijmen and Hinton, Geoffrey},
  howpublished = {COURSERA: Neural Networks for Machine Learning},
  year         = {2012}
}

@inproceedings{alonso2024diamond,
  title     = {Diffusion for World Modeling: Visual Details Matter in {Atari}},
  author    = {Alonso, Eloi and Jelley, Adam and Micheli, Vincent and Kanervisto, Anssi and Storkey, Amos and Pearce, Tim and Fleuret, Fran{\c{c}}ois},
  booktitle = {Advances in Neural Information Processing Systems (NeurIPS)},
  year      = {2024}
}

@inproceedings{bruce2024genie,
  title     = {Genie: Generative Interactive Environments},
  author    = {Bruce, Jake and Dennis, Michael and Edwards, Ashley and Parker-Holder, Jack and Shi, Yuge and Hughes, Edward and Lai, Matthew and Mavalankar, Aditi and Steigerwald, Richie and Apps, Chris and Aytar, Yusuf and Bechtle, Sarah and Behbahani, Feryal and Chan, Stephanie and Heess, Nicolas and Gonzalez, Lucy and Osindero, Simon and Ozair, Sherjil and Reed, Scott and Zhang, Jingwei and Zolna, Konrad and Clune, Jeff and de Freitas, Nando and Singh, Satinder and Rockt{\"a}schel, Tim},
  booktitle = {International Conference on Machine Learning (ICML)},
  year      = {2024}
}

@inproceedings{zhang2023storm,
  title     = {{STORM}: Efficient Stochastic Transformer based World Models for Reinforcement Learning},
  author    = {Zhang, Weipu and Wang, Gang and Sun, Jian and Yuan, Yetian and Huang, Gao},
  booktitle = {Advances in Neural Information Processing Systems (NeurIPS)},
  year      = {2023}
}

@article{ding2024worldmodelsurvey,
  title   = {Understanding World or Predicting Future? A Comprehensive Survey of World Models},
  author  = {Ding, Jingtao and Zhang, Yunke and Shang, Yu and Zhang, Yuheng and Zong, Zefang and Feng, Jie and Yuan, Yuan and Su, Hongyuan and Li, Nian and Sukiennik, Nicholas and Xu, Fengli and Li, Yong},
  journal = {arXiv preprint arXiv:2411.14499},
  year    = {2024}
}

@inproceedings{bai2025rat,
  title={Rat: Adversarial attacks on deep reinforcement agents for targeted behaviors},
  author={Bai, Fengshuo and Liu, Runze and Du, Yali and Wen, Ying and Yang, Yaodong},
  booktitle={Proceedings of the AAAI Conference on Artificial Intelligence},
  volume={39},
  pages={15453--15461},
  year={2025}
}

@article{tassa2018dmc,
  title={Deepmind control suite},
  author={Tassa, Yuval and Doron, Yotam and Muldal, Alistair and Erez, Tom and Li, Yazhe and Casas, Diego de Las and Budden, David and Abdolmaleki, Abbas and Merel, Josh and Lefrancq, Andrew and others},
  journal={arXiv preprint arXiv:1801.00690},
  year={2018}
}

@article{bellemare2013ale,
  title   = {The Arcade Learning Environment: An Evaluation Platform for General Agents},
  author  = {Bellemare, Marc G. and Naddaf, Yavar and Veness, Joel and Bowling, Michael},
  journal = {Journal of Artificial Intelligence Research},
  volume  = {47},
  pages   = {253--279},
  year    = {2013}
}

@article{mnih2015human,
  title   = {Human-Level Control through Deep Reinforcement Learning},
  author  = {Mnih, Volodymyr and Kavukcuoglu, Koray and Silver, David and Rusu, Andrei A. and Veness, Joel and Bellemare, Marc G. and Graves, Alex and Riedmiller, Martin and Fidjeland, Andreas K. and Ostrovski, Georg and Petersen, Stig and Beattie, Charles and Sadik, Amir and Antonoglou, Ioannis and King, Helen and Kumaran, Dharshan and Wierstra, Daan and Legg, Shane and Hassabis, Demis},
  journal = {Nature},
  volume  = {518},
  number  = {7540},
  pages   = {529--533},
  year    = {2015}
}

@article{hafner2021crafter,
  title   = {Benchmarking the Spectrum of Agent Capabilities},
  author  = {Hafner, Danijar},
  journal = {arXiv preprint arXiv:2109.06780},
  year    = {2021}
}

@article{wang2004ssim,
  title   = {Image Quality Assessment: From Error Visibility to Structural Similarity},
  author  = {Wang, Zhou and Bovik, Alan C. and Sheikh, Hamid R. and Simoncelli, Eero P.},
  journal = {IEEE Transactions on Image Processing},
  volume  = {13},
  number  = {4},
  pages   = {600--612},
  year    = {2004}
}

@article{kullback1951information,
  title   = {On Information and Sufficiency},
  author  = {Kullback, Solomon and Leibler, Richard A.},
  journal = {Annals of Mathematical Statistics},
  volume  = {22},
  number  = {1},
  pages   = {79--86},
  year    = {1951}
}

@inproceedings{liang2020efficient,
title={Efficient adversarial attacks for visual object tracking},
author={Liang, Siyuan and Wei, Xingxing and Yao, Siyuan and Cao, Xiaochun},
booktitle={Computer Vision--ECCV 2020: 16th European Conference, Glasgow, UK, August 23--28, 2020, Proceedings, Part XXVI 16},
year={2020},
}

@article{wei2018transferable,
title={Transferable adversarial attacks for image and video object detection},
author={Wei, Xingxing and Liang, Siyuan and Chen, Ning and Cao, Xiaochun},
journal={arXiv preprint arXiv:1811.12641},
year={2018}
}

@inproceedings{liang2021generate,
title={Generate more imperceptible adversarial examples for object detection},
author={Liang, Siyuan and Wei, Xingxing and Cao, Xiaochun},
booktitle={ICML 2021 Workshop on Adversarial Machine Learning},
year={2021}
}

@article{liang2022parallel,
title={Parallel rectangle flip attack: A query-based black-box attack against object detection},
author={Liang, Siyuan and Wu, Baoyuan and Fan, Yanbo and Wei, Xingxing and Cao, Xiaochun},
journal={arXiv preprint arXiv:2201.08970},
year={2022}
}

@inproceedings{liang2022large,
title={A large-scale multiple-objective method for black-box attack against object detection},
author={Liang, Siyuan and Li, Longkang and Fan, Yanbo and Jia, Xiaojun and Li, Jingzhi and Wu, Baoyuan and Cao, Xiaochun},
booktitle={European Conference on Computer Vision},
year={2022},
}

@inproceedings{liu2023x,
  title = {{\{X-Adv\}}: Physical Adversarial Object Attacks Against X-Ray Prohibited Item Detection},
  author = {Liu, Aishan and Guo, Jun and Wang, Jiakai and Liang, Siyuan and Tao, Renshuai and Zhou, Wenbo and Liu, Cong and Liu, Xianglong and Tao, Dacheng},
  booktitle = {32nd USENIX Security Symposium (USENIX Security 23)},
  year = {2023}
}

@article{wang2023diversifying,
title={Diversifying the High-level Features for better Adversarial Transferability},
author={Wang, Zhiyuan and Zhang, Zeliang and Liang, Siyuan and Wang, Xiaosen},
journal={arXiv preprint arXiv:2304.10136},
year={2023}
}

@article{liu2023improving,
title={Improving adversarial transferability by stable diffusion},
author={Liu, Jiayang and Zhu, Siyu and Liang, Siyuan and Zhang, Jie and Fang, Han and Zhang, Weiming and Chang, Ee-Chien},
journal={arXiv preprint arXiv:2311.11017},
year={2023}
}

@article{he2023sa,
title={SA-Attack: Improving Adversarial Transferability of Vision-Language Pre-training Models via Self-Augmentation},
author={He, Bangyan and Jia, Xiaojun and Liang, Siyuan and Lou, Tianrui and Liu, Yang and Cao, Xiaochun},
journal={arXiv preprint arXiv:2312.04913},
year={2023}
}

@article{zhang2024visual,
  title={Visual Adversarial Attack on Vision-Language Models for Autonomous Driving},
  author={Zhang, Tianyuan and Wang, Lu and Zhang, Xinwei and Zhang, Yitong and Jia, Boyi and Liang, Siyuan and Hu, Shengshan and Fu, Qiang and Liu, Aishan and Liu, Xianglong},
  journal={arXiv preprint arXiv:2411.18275},
  year={2024}
}

@article{wang2025black,
  title={Black-Box Adversarial Attack on Vision Language Models for Autonomous Driving},
  author={Wang, Lu and Zhang, Tianyuan and Qu, Yang and Liang, Siyuan and Chen, Yuwei and Liu, Aishan and Liu, Xianglong and Tao, Dacheng},
  journal={arXiv preprint arXiv:2501.13563},
  year={2025}
}

@article{kong2024patch,
  title={Patch is enough: naturalistic adversarial patch against vision-language pre-training models},
  author={Kong, Dehong and Liang, Siyuan and Zhu, Xiaopeng and Zhong, Yuansheng and Ren, Wenqi},
  journal={Visual Intelligence},
  volume={2},
  number={1},
  pages={1--10},
  year={2024},
  publisher={Springer}
}

@article{guo2026world,
  title={When World Models Dream Wrong: Physical-Conditioned Adversarial Attacks against World Models},
  author={Guo, Zhixiang and Liang, Siyuan and Balogh, Andras and Lunberry, Noah and Tu, Rong-Cheng and Jelasity, Mark and Tao, Dacheng},
  journal={arXiv preprint arXiv:2602.18739},
  year={2026}
}

@article{xu2026ctrlattack,
  title={CtrlAttack: A Unified Attack on World-Model Control in Diffusion Models},
  author={Xu, Shuhan and Liang, Siyuan and Zheng, Hongling and Luo, Yong and Hu, Han and Zhang, Lefei and Tao, Dacheng},
  journal={arXiv preprint arXiv:2603.13435},
  year={2026}
}

@article{lou2024hide,
title={Hide in Thicket: Generating Imperceptible and Rational Adversarial Perturbations on 3D Point Clouds},
author={Lou, Tianrui and Jia, Xiaojun and Gu, Jindong and Liu, Li and Liang, Siyuan and He, Bangyan and Cao, Xiaochun},
journal={arXiv preprint arXiv:2403.05247},
year={2024}
}

@inproceedings{wang2022bet,
  title={BET: black-box efficient testing for convolutional neural networks},
  author={Wang, Jialai and Qiu, Han and Rong, Yi and Ye, Hengkai and Li, Qi and Li, Zongpeng and Zhang, Chao},
  booktitle={Proceedings of the 31st ACM SIGSOFT International Symposium on Software Testing and Analysis},
  pages={164--175},
  year={2022}
}

@inproceedings{wang2025your,
  title={Your scale factors are my weapon: Targeted bit-flip attacks on vision transformers via scale factor manipulation},
  author={Wang, Jialai and Wu, Yuxiao and Xu, Weiye and Huang, Yating and Zhang, Chao and Li, Zongpeng and Xu, Mingwei and Liang, Zhenkai},
  booktitle={Proceedings of the Computer Vision and Pattern Recognition Conference},
  pages={20103--20112},
  year={2025}
}

@article{liu2024OPT,
  author={Liu, Shunyu and Song, Jie and Zhou, Yihe and Yu, Na and Chen, Kaixuan and Feng, Zunlei and Song, Mingli},
  journal={IEEE Transactions on Pattern Analysis and Machine Intelligence}, 
  title={Interaction Pattern Disentangling for Multi-Agent Reinforcement Learning}, 
  year={2024},
  volume={46},
  number={12},
  pages={8157-8172},
}

@inproceedings{liu2023contrastive,
  title={Contrastive identity-aware learning for multi-agent value decomposition},
  author={Liu, Shunyu and Zhou, Yihe and Song, Jie and Zheng, Tongya and Chen, Kaixuan and Zhu, Tongtian and Feng, Zunlei and Song, Mingli},
  booktitle={Proceedings of the AAAI conference on artificial intelligence},
  volume={37},
  number={10},
  pages={11595--11603},
  year={2023}
}

@article{jing2024deep,
  title={Deep graph mating},
  author={Jing, Yongcheng and Hong, Seok-Hee and Tao, Dacheng},
  journal={Advances in Neural Information Processing Systems},
  volume={37},
  pages={9753--9772},
  year={2024}
}

@inproceedings{zhao2021multi,
  title={Multi-attentional deepfake detection},
  author={Zhao, Hanqing and Zhou, Wenbo and Chen, Dongdong and Wei, Tianyi and Zhang, Weiming and Yu, Nenghai},
  booktitle={Proceedings of the IEEE/CVF conference on computer vision and pattern recognition},
  pages={2185--2194},
  year={2021}
}
